\documentclass[lettersize,journal]{IEEEtran}
\usepackage{amsmath,amsfonts}
\usepackage{algorithmic}
\usepackage{algorithm}
\usepackage{array}
\usepackage[caption=false,font=normalsize,labelfont=sf,textfont=sf]{subfig}
\usepackage{textcomp}
\usepackage{stfloats}
\usepackage{url}
\usepackage{verbatim}
\usepackage{graphicx}
\usepackage[numbers]{natbib}
\usepackage{multirow}
\usepackage{amssymb}
\usepackage{color} 

\usepackage[colorlinks,
            linkcolor=black,
            anchorcolor=white,
            citecolor=black]{hyperref}

\begin{document}

\title{Towards Understanding and Boosting Adversarial Transferability from a Distribution Perspective}

\author{Yao Zhu, Yuefeng Chen, Xiaodan Li, Kejiang Chen, Yuan He, Xiang Tian, Bolun Zheng, Yaowu Chen,

Qingming Huang,~\IEEEmembership{Fellow,IEEE} 
\thanks{Yao Zhu is with the Zhejiang University, Hangzhou,
China, 310013.\protect\\
(E-mail: ee$\_$zhuy$@$zju.edu.cn)}
\thanks{Yuefeng Chen, Xiaodan Li and Yuan He are with the Security Department of Alibaba Group. (E-mail: 
yuefeng.chenyf@alibaba-inc.com,fiona.lxd@alibaba-inc.com, heyuan.hy@alibaba-inc.com)}
\thanks{Kejiang Chen is with the CAS Key Laboratory of Electro-Magnetic Space Information, University of Science and Technology of China. (E-mail:chenkj@ustc.edu.cn)}
\thanks{Bolun Zheng is with the Hangzhou Dianzi University and also with Zhejiang Provincial Key Laboratory for Network Multimedia Technologies. (E-mail: blzheng@hdu.edu.cn)}
\thanks{Xiang Tian is with the Zhejiang University and also with Zhejiang Provincial Key Laboratory for Network Multimedia Technologies. (E-mail: xiang.t@163.com)}
\thanks{Yaowu Chen is with the Zhejiang University and also with Zhejiang University Embedded System Engineering Research Center, Ministry of Education of China. (E-mail: cyw@mail.bme.zju.edu.cn)}
\thanks{Qingming Huang is with the University of Chinese Academy of Sciences and also with the Key Laboratory of Intelligent Information Processing, Chinese Academy of Sciences. (E-mail: qmhuang@ucas.ac.cn)}
\thanks{Corresponding authors: Xiang Tian, Bolun Zheng.}

}



\maketitle
{
\begin{abstract}

Transferable adversarial attacks against Deep neural networks (DNNs) have received broad attention in recent years. An adversarial example can be crafted by a surrogate model and then attack the unknown target model successfully, which brings a severe threat to DNNs. The exact underlying reasons for the transferability are still not completely understood. Previous work mostly explores the causes from the model perspective, e.g., decision boundary, model architecture, and model capacity. adversarial attacks against Deep neural networks (DNNs) have received broad attention in recent years. An adversarial example can be crafted by a surrogate model and then attack the unknown target model successfully, which brings a severe threat to DNNs. The exact underlying reasons for the transferability are still not completely understood. Previous work mostly explores the causes from the model perspective, e.g., decision boundary, model architecture, and model capacity.
Here, we investigate the transferability from the data distribution perspective and hypothesize that pushing the image away from its original distribution can enhance the adversarial transferability.  
To be specific, moving the image out of its original distribution makes different models hardly classify the image correctly, which benefits the untargeted attack, and dragging the image into the target distribution misleads the models to classify the image as the target class, which benefits the targeted attack.
Towards this end, we propose a novel method that crafts adversarial examples by manipulating the distribution of the image. We conduct comprehensive transferable attacks against multiple DNNs to demonstrate the effectiveness of the proposed method. Our method can significantly improve the transferability of the crafted attacks and achieves state-of-the-art performance in both untargeted and targeted scenarios, surpassing the previous best method by up to 40$\%$ in some cases. In summary, our work provides new insight into studying adversarial transferability and provides a strong counterpart for future research on adversarial defense \footnote{The code will be available at \href{https://github.com/alibaba/easyrobust}{this https URL}}.

\end{abstract}
}

\begin{IEEEkeywords}
Adversarial transferability, adversarial attack, black-box attack.
\end{IEEEkeywords}

\section{Introduction}

\IEEEPARstart{D}{eep} neural networks (DNNs) have achieved great success in many fields, such as face recognition ~\cite{TIP_face_attack,facerec_tifs2,facerec_tifs3}, autonomous driving~\cite{autonomous_tifs,autonomous_cvprwork,autonomous_cvprwork2}, and speaker verification~\cite{speaker_tifs,speaker_tifs2,speaker_tifs3}. However,~\citet{szegedy2013intriguing} and \cite{goodfellow2014explaining} found that the imperceptible adversarial examples can be catastrophic for the DNNs.
{ Even worse, researchers found that adversarial examples can even transfer between the models with different architectures and parameters \cite{goodfellow2014explaining,liu2017delving}, which allows the attackers to attack unknown target models using adversarial examples generated by the surrogate models. 
Adversarial transferability has received more and more attention in recent years. On the one hand, such a phenomenon raises severe concerns about the security and safety of DNNs when deployed in real-world scenarios from both academia and industry. On the other hand, exploring the adversarial transferability would benefit many aspects, including understanding the deep learning models, developing stronger defenses and robust models, and evaluating the vulnerability of the modern DNNs \cite{goodfellow2014explaining}. }

{
Various understandings of adversarial transferability have been proposed in the past years and led to effective adversarial attacks. Most works explain such transferability from a model perspective, claiming that the decision boundary \cite{liu2017delving}, model architecture \cite{wu2020skip,su2019robustness}, and the test accuracy \cite{wu2020towardunderstanding,yang2021trs} of the surrogate model have a significant influence on the adversarial transferability.
These understandings of adversarial transferability from a model perspective motivate various methods to improve adversarial transferability by investigating models' properties.
Some works introduce data augmentation \cite{xie2019improvingDI,lin2020nesterov_NI-FGSM,admix} into the generation of adversarial examples or training generators \cite{naseer2019cross,naseer2021generatingTTP} to perform attacks to reduce the reliance on the decision boundary of the surrogate classifier. \citet{wu2020skip} propose to modify the architecture of the model to enhance the adversarial transferability and \citet{huang2020enhancingILA} propose to fine-tune the adversarial examples using the mid features of the surrogate model.
Though these methods are effective in untargeted scenarios, their performance is highly limited in targeted attack scenarios.
 }

 {
 To fully understand adversarial transferability, especially in targeted attack scenarios, we propose a novel perspective from the data distribution. 
 Recall the classical assumption in machine learning that the validation data that are independent and identically distributed with the training dataset can be classified correctly by different models, while the out-of-distribution examples can cause difficulty for models to classify \cite{wald2021calibrationOOD,nagarajan2020understandingOOD}.
 Our hypothesis is also built on such assumption. To be specific, we denote the distribution of the training dataset as $p_D(\boldsymbol{x}|y)$, where $y$ represents the class label, and $\boldsymbol{x}$ represents the image. Different models tend to predict the validation data that are identically distributed with $p_D(\boldsymbol{x}|y)$ as $y$ and can hardly classify the data that is not identically distributed with $p_D(\boldsymbol{x}|y)$ as $y$. Therefore, moving the image out of its original distribution causes difficulties for different models to classify this out-of-distribution example, which can enhance the transferability of the untargeted attack. Dragging the image into the target distribution misleads different models to classify the image as the target class, which can enhance the transferability of the targeted attack. }

\begin{figure}[htbp]
\centering
\includegraphics[width=0.495\textwidth]{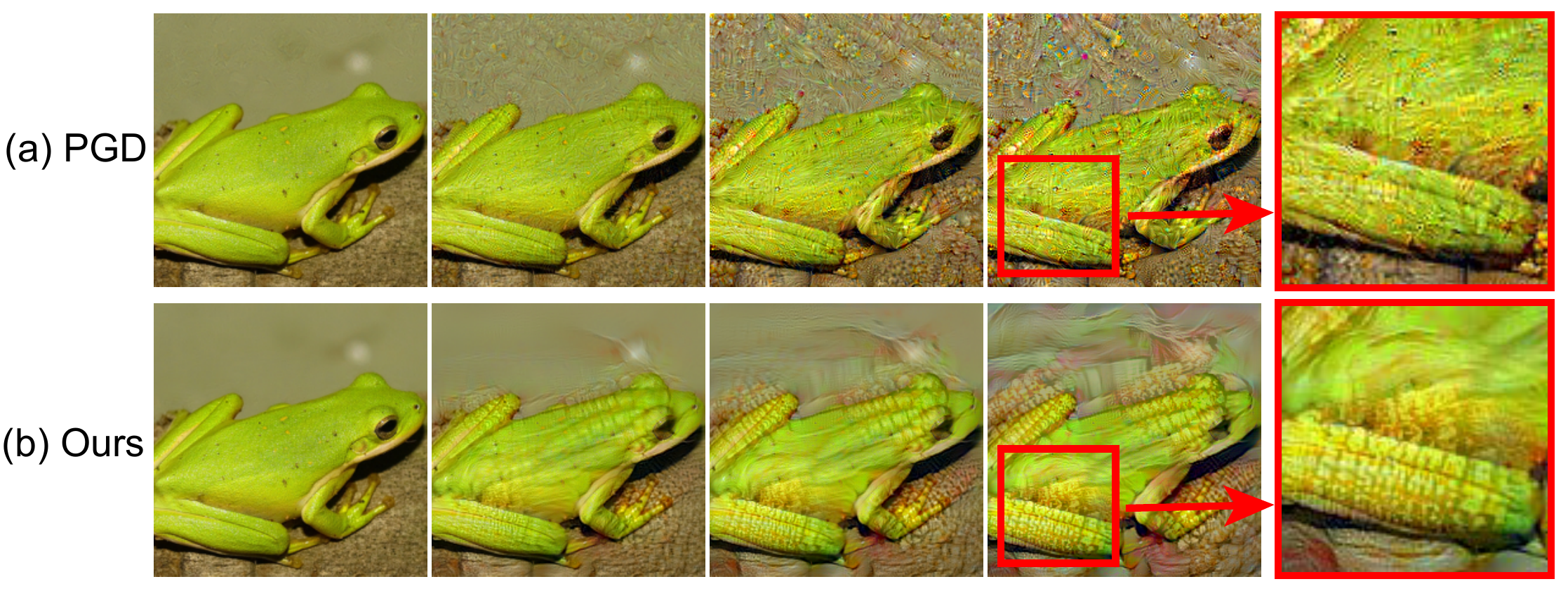}
\caption{Comparison between the targeted adversarial example generated by the normal PGD attack and our distribution-relevant attack. We set the maximum allowable adversarial perturbation as $\epsilon$ = 32/255 with respect to a pixel value in [0, 1] for visibility. The leftmost column of each figure shows the original image (tree frog), while the other column shows the adversarial images (target class: corn) under different attack strengths. The rightmost column amplifies the square patch in the adversarial examples.}
\label{img:corn}
\end{figure}

{
Towards this end, we propose a method named  \textbf{D}istribution-\textbf{R}elevant \textbf{A}ttack (\textbf{DRA}) to demonstrate our hypothesis. We attempt to push the input image away from its original distribution to generate transferable adversarial examples. However, as we do not have access to the ground truth data distribution, it is technically challenging to push the images away from its original distribution directly. 

We borrow the idea from the score-matching generative models \cite{scorematching,song2019sliced,score-based-gi,energyrobust}, which propose to estimate the gradient of the ground truth data distribution $\nabla_{\boldsymbol{x}} \log p_D(\boldsymbol{x}|y)$ and generate the image of the certain distribution iteratively using the estimated gradient of the ground truth data distribution through Langevin dynamics \cite{scorematching,score-based-gi}. 
Previous attacks iteratively minimize (maximize) the conditional density of the model $p_\theta(y|\boldsymbol{x})$ along the gradient of the conditional density of the model  $\nabla_{\boldsymbol{x}} \log p_\theta(y|\boldsymbol{x})$ to perform the untargeted (targeted) attacks. 
Thus, to estimate the gradient of the ground truth data distribution in the transfer attack scenarios, we fine-tune the surrogate classifier to match the gradient of the conditional density of the model and the gradient of ground truth data distribution.
Thereby, the gradient of the fine-tuned model can approximate the gradient of the ground truth data distribution and the process of generating the adversarial examples with the gradient of our fine-tuned models can approximate the process of the Langevin dynamics, which enables us to manipulate the distribution of the image. We name the attack that uses our fine-tuned models to push the image away from the original distribution while generating the adversarial examples \textbf{D}istribution-\textbf{R}elevant \textbf{A}ttack (\textbf{DRA}). What's more, \textbf{DRA} is compatible with existing transfer attacks and can greatly improve the performance of these attacks.

 Visually, targeted adversarial perturbation generated by our method, which can drag the image into the target distribution, reflects vivid semantic features of the target class (See Fig.\ref{img:corn}: Turning the tree frog to corn). In Fig.\ref{img:OOD_eval}, we use the out-of-distribution (OOD) detection method Energy \cite{liu2020energy} to evaluate that our \textbf{DRA} can indeed move the image out of its original distribution, performing better than the normal PGD attack.

\begin{figure}[htbp]
\centering
\includegraphics[width=0.495\textwidth]{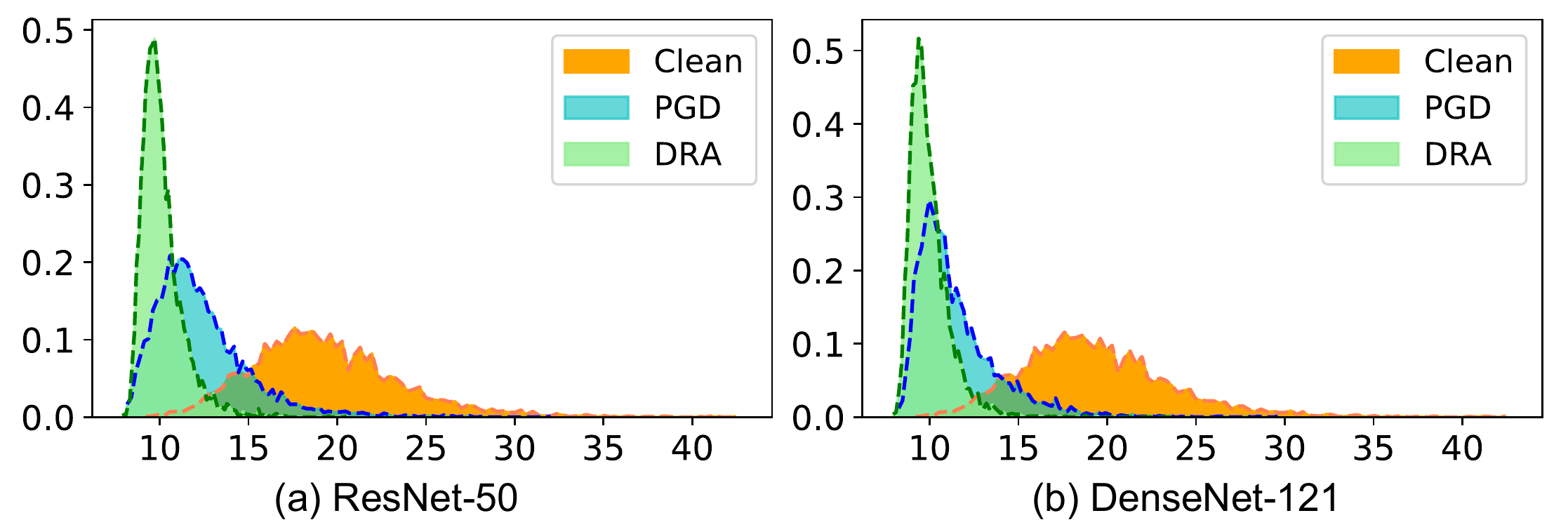}
\caption{The distribution of the Energy OOD scores \cite{liu2020energy} for the in-distribution images (ImageNet) and the untargeted adversarial examples generated by the original PGD attack and our \textbf{DRA} on ResNet-50 (a) and DenseNet-121 (b). Examples with lower OOD scores than the clean images are regarded as the out-of-distribution (OOD) examples and examples with higher energy scores are regarded as the in-distribution (ID) examples. }
\label{img:OOD_eval}
\end{figure}

}
 
 {

 We have conducted extensive evaluations and established state-of-the-art performance in both untargeted and targeted attack scenarios, improving the targeted attack success rate by up to 40$\%$ in some cases. This work provides new insight into the understanding of adversarial transferability from a data distribution perspective and provides a strong counterpart for future research on defense.
 }

The main contributions of our paper are summarized as follows:
 \begin{itemize}

  \item[$\bullet$] We provide a new understanding of adversarial transferability from the perspective of data distribution, advocating that adversarial transferability can be enhanced by pushing the images away from its original distribution.
 
 \item[$\bullet$] We introduce a method to match the gradient of the model and the gradient of the data distribution, which enables us to push the image away from its original distribution using the gradient of the model. 

 \item[$\bullet$] Extensive experiments demonstrate that our \textbf{DRA} outperforms state-of-the-art approaches a lot in both untargeted and targeted attack scenarios (even up to 40\% in most cases).
 
 \end{itemize}
 
The rest of the paper is organized as follows. Section II summarizes the literature related to adversarial attacks. {In Section III, we firstly present some preliminaries and the motivation of our method.} 
Then we introduce the optimization of the distance between the gradient of the model and the gradient of the data distribution.
After that, we propose the Algorithm of our \textbf{DRA}, which fine-tunes the original surrogate model and use the fine-tuned model to generate adversarial examples. In Section IV, we firstly conduct untargeted transfer attack experiments to demonstrate the superiority of \textbf{DRA} against various target models, including both normal models and secured models.
Then, we evaluate the effectiveness of \textbf{DRA} in the targeted attack scenario which is more difficult than the untargeted scenario. {Section V provides some discussion to further understand our method.} Section VI gives some conclusive results.


\section{Related Work}
 In this section, we briefly review the literature related to adversarial attacks.

 Deep neural networks (DNNs) obtained by normal training are vulnerable to adversarial examples~\cite{szegedy2013intriguing}. This phenomenon has drawn wide concern and affected the deployment of DNNs in many safety-critical fields, such as face recognition~\cite{facetrans,komkov2019advhat,facecrossdomain},  medical diagnosis~\cite{ma2020understanding,attackmedical}, speaker recognition~\cite{speakerattack,speakattack2}, and autonomous driving~\citep{morgulis2019fooling,attackdriving}.
 According to the access rights to the target model, adversarial attacks can be classified as white-box attacks and black-box attacks.
 
 White-box attacks assume that the attacker can completely access the structure and parameters of the target model. 
 Typical examples of white-box attacks are FGSM~\cite{goodfellow2014explaining}, BIM~\cite{BIM}, PGD~\cite{madry2019kpgd}, DeepFool~\cite{Deepfool}, JSM~\cite{JSM}, and CW~\cite{CW}. 
 The widely used PGD attack builds an adversarial example by performing multi-step gradient updating along the direction of the gradient at each pixel and projecting the perturbation into the specified range.
 
 Black-box attacks assume that the attacker only knows the output of the target model (prediction or confidence), including query-based attacks and transfer attacks. Query-based attacks perform the black-box attack by estimating gradient with queries to the target model~\cite{papernot2017practical_black,Su_onepixel_black,faceattack_query,yang2020learning_black}. In this paper, we assume that the attacker can only generate adversarial perturbation using a surrogate model without any queries on the target model. This method is much more efficient and is relatively harder to detect by the target system than query-based attacks. Thus, many existing works focused on leveraging the transferability of adversarial examples.
 
 \textbf{Iterative Methods:} ~\citet{liu2017delving} show that adversarial attacks can be transferred between different models. Iterative methods attack a surrogate model and update the perturbation iteratively using gradient information~\cite{dong2018boostingMI,xie2019improvingDI,lin2020nesterov_NI-FGSM,dong2019TI,wu2020skip,zhangIR,guo2020LinBP,zhao2021_simplicity}. The iterative attack methods such as BIM~\cite{BIM} and PGD~\cite{madry2019kpgd} could achieve good performance in white-box attack scenarios, but they often suffer from low transferability. Recently, many methods have been proposed to improve the adversarial transferability of the iterative methods.
 
 Some methods suggest stabilizing update directions for the iterative algorithms.
 ~\citet{dong2018boostingMI} propose to improve the adversarial transferability by integrating the momentum of gradients into the update of perturbation.~\citet{lin2020nesterov_NI-FGSM} adapt Nesterov accelerated gradient into the update of perturbation to enhance the adversarial transferability. 
 Data augmentation, which plays an important role in improving model generalization and mitigating over-fitting, also contributes to adversarial transferability.~\citet{xie2019improvingDI} suggest applying random transformations (resizing and padding) to the input images at each iteration during attacking.~\citet{lin2020nesterov_NI-FGSM} take the scale copies of the input images into attack in order to mitigate overfitting on the surrogate model.~\citet{admix} propose Admix, which attacks the input image admixed with a group of images randomly sampled from other categories.~\citet{zhangIR} propose a loss to decrease interactions between perturbation units during attacking. 
 There are also some model-specific methods to improve adversarial transferability.
 ~\citet{huang2020enhancingILA} fine-tune the adversarial examples by increasing perturbation on a pre-specified layer.~\citet{wu2020skip} propose to reduce the gradients from the residual modules and pay attention to the architectural vulnerability of DNNs.~\citet{guo2020LinBP} show that properly increasing the linearity of DNNs can enhance adversarial transferability. 
 
 These methods perform well in untargeted attacks, but their performance degrades severely in targeted attacks.
 
 \textbf{Generative Methods:} 
 The methods generating the targeted adversarial perturbations by generative models always perform better than iterative methods at the expense of training the same number of generative models as the labels~\cite{kanth2021learning,poursaeed2018generative,nag-cvpr-2018,naseer2019cross,naseer2021generatingTTP}.~\citet{poursaeed2018generative} propose to train the generative model against the surrogate classifier via cross-entropy loss.~\citet{naseer2019cross} show that the relativistic cross-entropy loss can improve the performance of the generative model.~\citet{naseer2021generatingTTP} propose to match the ‘distribution’ of perturbed images with that of the target class within latent space of the surrogate classifier in generative training so as to reduce the reliance on class-boundary information from the surrogate classifier. This method can successfully imprint the features of the target distribution to the image and achieves satisfactory targeted attack performance but needs to train generators for every class which is nontrivial on large-scale datasets. 

 Compared with the existing iterative methods, our \textbf{DRA} pays attention to the distribution-relevant information in the surrogate model rather than improving the iterative algorithm or performing data augmentation. \textbf{DRA} overcomes the low transferability of iterative attacks in targeted attack scenarios. The generative methods aim to learn the distribution of the adversarial perturbation, while our \textbf{DRA} focuses on the ground truth distribution. As we show in our experiments, our \textbf{DRA} greatly improves the adversarial transferability and surpasses existing methods in both untargeted and targeted scenarios.

\section{Method}

{
\subsection{Preliminary}

 Given a surrogate classifier $f_{\theta}$ parameterized by $\theta$, and image $\boldsymbol{x}$, label $y$, total possible classes $n$, then $f_{\theta}(\boldsymbol{x})[k]$ represents the $k^{th}$ output of the last layer. 

 The conditional density $p_{\theta}(y|{\boldsymbol{x}})$ can be expressed as:
 \begin{equation}
 p_{\theta}(y|\boldsymbol{x}) = \frac{\exp({f_{\theta}(\boldsymbol{x})[y]})}{\sum_{k=1}^{n} \exp({f_{\theta}(\boldsymbol{x})[k]})}.
 \label{p(ygx)}    
 \end{equation}
 
 The adversarial perturbation is usually based on the gradient of the classification loss $\mathcal{L}$. The untargeted attack aims to minimize the conditional density $p_\theta(y|\boldsymbol{x})$ and can be expressed as~\cite{goodfellow2014explaining,madry2019kpgd}:
 \begin{equation}
 \boldsymbol{x}' = \boldsymbol{x}+\nabla_{\boldsymbol{x}} \mathcal{L}(f_{\theta}(\boldsymbol{x}),y) = \boldsymbol{x}-\nabla_{\boldsymbol{\boldsymbol{x}}} \log p_\theta(y|\boldsymbol{x}),
 \label{untarget_attack}    
 \end{equation} 
 where $\boldsymbol{x}'$ means the adversarial example of the image $\boldsymbol{x}$. The targeted attack aims to maximize the conditional density $p_\theta(y_{target}|\boldsymbol{x})$ and can be expressed as:
 \begin{equation}
 \boldsymbol{x}' = \boldsymbol{x}-\nabla_{\boldsymbol{x}} \mathcal{L}(f_{\theta}(\boldsymbol{x}),y_{target}) = \boldsymbol{x}+\nabla_{\boldsymbol{\boldsymbol{x}}} \log p_\theta(y_{target}|\boldsymbol{x}).
 \label{target_attack}    
 \end{equation} 
 
 The goal of the transfer attack is to mislead the target model using the adversarial examples generated by the surrogate model.}
 
{ 
\subsection{Motivation}


The existing transfer attacks iteratively minimize $p_\theta(y_{label}|\boldsymbol{x})$ (untargeted attack) or maximize $p_\theta(y_{target}|\boldsymbol{x})$ (targeted attack) of the surrogate model to generate adversarial examples and then use these adversarial examples to attack the target models.
However, the existing transfer attacks can hardly perform targeted attacks successfully and lack the explanation why minimizing $p_\theta(y|\boldsymbol{x})$ of the surrogate model can also fool the target model with different model parameters and architectures from the surrogate model.


In this paper, we propose to understand and improve the adversarial transferability from a data distribution perspective, which builds on the classical assumption in machine learning methods \cite{wald2021calibrationOOD,nagarajan2020understandingOOD} that the deep models can properly classify the validation data that is independent and identically distributed with the training dataset but can hardly classify the out-of-distribution examples. 
Specifically, the models tend to predict the label of the image that is identically distributed with $p_D(\boldsymbol{x}|y)$ as $y$, but can not handle the out of distribution images properly.
We hypothesize that moving the image out of its original distribution can achieve high untargeted adversarial transferability and dragging the image into the target distribution $p_D(\boldsymbol{x}|y_{target})$ can achieve high targeted adversarial transferability.
The challenge comes from how to push the image away from its original distribution as we don't have access to the ground truth class-conditional data distribution $p_D(\boldsymbol{x}|y)$.


We borrow the idea from the score-matching generative models \cite{scorematching,song2019sliced,score-based-gi} which propose to estimate the gradient of the ground truth data distribution and then move the initial image from its original distribution $p_D(\boldsymbol{x}|y_0)$ to the target distribution $p_D(\boldsymbol{x}|y)$ iteratively through the Stochastic Gradient Langevin Dynamics (SGLD) \cite{energyrobust,lecun2006,SGLD}:
\begin{equation}
 \boldsymbol{x}_{t} = \boldsymbol{x}_{t-1} + \alpha\cdot\nabla_{\boldsymbol{x}_{t-1}} \log p_D(\boldsymbol{x}_{t-1}|y) + \sqrt{2\alpha}\cdot\epsilon.
 \label{sgld3_}
 \end{equation}
The $\epsilon \sim \mathcal{N}(0,I)$ and $\alpha$ is a fixed step size. When $\alpha \rightarrow 0$ and $T\rightarrow \infty$, $\boldsymbol{x}_T$ is exactly an sample from $p_D(\boldsymbol{x}|y)$. Updating the SGLD process along the opposite direction of $\nabla_{\boldsymbol{x}} \log p_D(\boldsymbol{x}|y)$ can move the image away from the distribution $p_D(\boldsymbol{x}|y)$. Based on the above {reasoning}, the gradient of the data distribution can be used to manipulate the distribution of the input via iterative methods.

In this paper, we propose to match the gradient of the log conditional density $\nabla_{\boldsymbol{x}} \log p_\theta (y|\boldsymbol{x})$ (the direction of the normal adversarial attack) and the gradient of the log ground truth class-conditional data distribution $\nabla_{\boldsymbol{x}} \log p_D (\boldsymbol{x}|y)$. In this way, the adversarial attack can approximate the direction of the gradient of the ground truth class-conditional data distribution. 

To be specific, if $\nabla_{\boldsymbol{x}} \log p_\theta (y|\boldsymbol{x})$ matches $\nabla_{\boldsymbol{x}} \log p_D (\boldsymbol{x}|y)$ well, the untargeted attack $x' = x - \eta \cdot \nabla_{\boldsymbol{\boldsymbol{x}}} \log p_\theta(y_{label}|\boldsymbol{x})$ can be {regarded} as an approximation of the opposite process of SGLD sampling $\boldsymbol{x}_{t} = \boldsymbol{x}_{t-1} - \alpha\cdot\nabla_{\boldsymbol{x}_{t-1}} \log p_D(\boldsymbol{x}_{t-1}|y_{label})$, which moves the images out of its original distribution $p_D(\boldsymbol{x}|y_{label})$.
Fig.\ref{img:OOD_eval} shows that our untargeted attack indeed moves the image out of its original distribution, which causes difficulties for different models to classify this image.

Similarly, if $\nabla_{\boldsymbol{x}} \log p_\theta (y|\boldsymbol{x})$ matches $\nabla_{\boldsymbol{x}} \log p_D (\boldsymbol{x}|y)$ well, our targeted attack $x' = x + \eta \cdot \nabla_{\boldsymbol{\boldsymbol{x}}} \log p_\theta(y_{target}|\boldsymbol{x})$ can be {regarded} as an approximation of the process of SGLD sampling $\boldsymbol{x}_{t} = \boldsymbol{x}_{t-1} + \alpha\cdot\nabla_{\boldsymbol{x}_{t-1}} \log p_D(\boldsymbol{x}_{t-1}|y_{target})$, which drags the image to the target distribution $p_D(\boldsymbol{x}|y_{target})$. Fig.\ref{img:corn} shows that our method can imprint the features of the target distribution to the image and semantically change the tree-frog to corn, which can mislead the models to classify the image as the target class. Compared with the existing transfer attacks, our method aims to intrinsically manipulate the distribution of the image rather than just minimizing or maximizing the classification loss.

In the next subsection, we provide an appealingly simple and generic technique to match the $\nabla_{\boldsymbol{x}} \log p_\theta (y|\boldsymbol{x})$ and $\nabla_{\boldsymbol{x}} \log p_D (\boldsymbol{x}|y)$. The last subsection instructs how to generate high transferable adversarial examples with our method.

}

\subsection{Decreasing the Distance Between Gradients}
{
 In this section, we propose a novel method to decrease the distance between the gradients, which enables us to use the gradient of the model to estimate the gradient of the ground truth data distribution. In this way, adversarial attack can push the image away from its original distribution through Langevin Dynamics (Eq.(\ref{sgld3_})).
 }
 
 We define the \textbf{D}istance between the gradient of log \textbf{C}onditional density and the gradient of log \textbf{G}round truth class-conditional data distribution ({\rm DCG}) as :

\begin{small}
\begin{equation}
\begin{aligned}
     &\mathop{\rm{DCG}}\triangleq {\mathbb{E}_{p_D(y)}}\mathbb{E}_{p_D(\boldsymbol{x}|y)}\Vert \nabla_{\boldsymbol{x}}\log p_{\theta}(y|\boldsymbol{x})-\nabla_{\boldsymbol{x}}\log p_D({\boldsymbol{x}}|y) \Vert_2^2\\
    &=\int  \int   \Vert\nabla_{\boldsymbol{x}}\log p_{\theta}(y|\boldsymbol{x})-\nabla_{\boldsymbol{x}}\log p_D({\boldsymbol{x}}|y)\Vert_2^2 p_D(\boldsymbol{x}|y)p_D(y) d\boldsymbol{x}dy \\
    &=\int  \int   \Vert\nabla_{\boldsymbol{x}}\log p_D({\boldsymbol{x}}|y)\Vert_2^2 p_D(\boldsymbol{x}|y) p_D(y) d\boldsymbol{x} dy\\
    &+ \int  \int   \Vert\nabla_{\boldsymbol{x}}\log p_{\theta}(y|\boldsymbol{x})\Vert_2^2 p_D(\boldsymbol{x}|y) p_D(y) d\boldsymbol{x} dy \\
    &-2\int  \int   (\nabla_{\boldsymbol{x}}\log p_{\theta}(y|\boldsymbol{x})^{\rm{T}}\cdot\nabla_{\boldsymbol{x}}\log p_D({\boldsymbol{x}}|y)) p_D(\boldsymbol{x}|y) p_D(y) d\boldsymbol{x} dy.
\label{ACGalignment_threeparts}
\end{aligned}
\end{equation}
\end{small}
We omit the integration domain here for simplicity. The first term is a constant which does not depend on the model's parameters $\theta$. The middle term can be expressed as $\mathbb{E}_{p_{D}(y)}\mathbb{E}_{p_{D}(\boldsymbol{x}|y)}  \left\|\nabla_{\boldsymbol{x}}\log p_{\theta}(y|\boldsymbol{x})\right\| _{2}^{2}$ is tractable since this term does not contain the unknown score of the ground truth distribution. The last term is not directly computable, because the score of the ground truth distribution $\nabla_{\boldsymbol{x}}\log p_D({\boldsymbol{x}}|y)$ is unknown. Score {matching} methods \cite{song2019sliced,score-based-gi,hyvarinen2005estimation} eliminate the score of the ground truth distribution using integration by parts. Inspired by these methods, we apply integration by parts to the last term in Eq.(\ref{ACGalignment_threeparts}) as:

\begin{footnotesize}
\begin{equation}
\begin{aligned}
     &\int_{-\infty}^{+\infty}\hspace{-0.2cm} p_D(y) dy \int_{\boldsymbol{x} \in \mathbb{R}^n}   (\nabla_{\boldsymbol{x}}\log p_{\theta}(y|\boldsymbol{x})^{\rm{T}}\cdot\nabla_{\boldsymbol{x}}\log p_D({\boldsymbol{x}}|y)) p_D(\boldsymbol{x}|y) d\boldsymbol{x} \\
    &\overset{\text{(I)}}= \int_{-\infty}^{+\infty}\hspace{-0.2cm} p_D(y) dy\int_{\boldsymbol{x} \in \mathbb{R}^n}   (\nabla_{\boldsymbol{x}}\log p_{\theta}(y|\boldsymbol{x})^{\rm{T}}\cdot\nabla_{\boldsymbol{x}} p_D({\boldsymbol{x}}|y))  d\boldsymbol{x} \\
    &\overset{\text{(II)}}= \int_{-\infty}^{+\infty}\hspace{-0.2cm} p_D(y) dy  \sum_{i=1}^{n} \int_{\boldsymbol{x} \in \mathbb{R}^n} \nabla_{{x_i}}\log p_{\theta}(y|\boldsymbol{x})\nabla_{{x_i}} p_D({\boldsymbol{x}}|y) d {\boldsymbol{x}}\\
    &\overset{\text{(III)}}=\hspace{-0.2cm}\int_{-\infty}^{+\infty}\hspace{-0.35cm} p_D(y) dy \sum_{i=1}^{n} \int_{\tilde{\boldsymbol{x}}_i \in \mathbb{R}^{n-1}} [\lim_{M\to\infty} p_D(\boldsymbol{x}|y) \nabla_{x_i}\log p_{\theta}(y|\boldsymbol{x})|_{-\boldsymbol{M}_i}^{+\boldsymbol{M}_i}]d{\tilde{\boldsymbol{x}}_i} \\
    &\quad - \int_{-\infty}^{+\infty}\hspace{-0.2cm} p_D(y) dy \sum_{i=1}^{n} \int_{\tilde{\boldsymbol{x}}_i \in \mathbb{R}^{n-1}} [\int_{-\infty}^{+\infty}\hspace{-0.2cm} p_D(\boldsymbol{x}|y) \nabla^2_{x_i} \log p_{\theta}(y|\boldsymbol{x}) d{x_i}]d{\tilde{\boldsymbol{x}}_i}\\
    &\overset{\text{(IV)}}=- \mathbb{E}_{p_D(y)}\mathbb{E}_{p_D({\boldsymbol{x}}|y)} \left[ {\rm{tr}} (\nabla^2_{\boldsymbol{x}} \log p_{\theta}(y|{\boldsymbol{x}}))\right],
\label{ACGalignment_third}
\end{aligned}
\end{equation}
\end{footnotesize}
where $\nabla^2_{\boldsymbol{x}}$ denotes the Hessian with respect to $\boldsymbol{x}$. ``+$\boldsymbol{M}_i$" represents the vector $[x_1,...,x_{i-1},+M,x_{i+1},...,x_n]$. ``-$\boldsymbol{M}_i$" represents the vector $[x_1,...,x_{i-1},-M,x_{i+1},...,x_n]$. $\boldsymbol{x}=[x_1,...,x_n]$ is an n-dimensional vector. $\tilde{\boldsymbol{x}}_i=[x_1,...,x_{i-1},x_{i+1},...,x_n]$. See Appendix for detailed derivation.

We use the formula: $\nabla_x \log f(x) = f(x)^{-1}\nabla_x f(x)$ for equality (I). 
{In equality (I), $\nabla_{\boldsymbol{x}}\log p_{\theta}(y|\boldsymbol{x})^{\rm{T}}$ and $\nabla_{\boldsymbol{x}}\log p_D({\boldsymbol{x}}|y)$ are n-dimensional vectors, and their product result is a scalar.}
We use the formula: $\boldsymbol{u}^T \cdot \boldsymbol{v} = \sum_{i=1}^{n} u_i v_i$ for equality (II), where n represents the dimension of the data. 
As for equality (III), we use the integration by parts formula (See Appendix for proof): 
\begin{footnotesize}
\begin{equation}
\begin{aligned}
\int_{-\infty}^{+\infty} \nabla_{{x_i}}f(\boldsymbol{x})\nabla_{{x_i}}g(\boldsymbol{x}) d{x_i}
&=  \lim_{M\to\infty}g({\boldsymbol{x}})\nabla_{{x_i}}f({\boldsymbol{x}})|_{-\boldsymbol{M}_i}^{+\boldsymbol{M}_i}\\
&-\int_{-\infty}^{+\infty}g(\boldsymbol{x})\nabla_{{x_i}}^2 f(\boldsymbol{x}) dx_i.
\label{integration_part}
\end{aligned}
\end{equation} 
\end{footnotesize}
The equality (IV) holds for that we assume $p_D({\boldsymbol{x}}|y) \rightarrow 0$ {when $||\boldsymbol{x}||_2 \rightarrow \infty$.}

Thus, substituting the results of integration by parts into Eq.(\ref{ACGalignment_threeparts}), the DCG loss can be reformulated as: 
 \begin{equation}
 \begin{aligned}
    \mathop{\rm{DCG}} &\triangleq \mathbb{E}_{p_D(y)}\mathbb{E}_{p_D(\boldsymbol{x}|y)}\Vert \nabla_{\boldsymbol{x}}\log p_{\theta}(y|\boldsymbol{x})-\nabla_{\boldsymbol{x}}\log p_D({\boldsymbol{x}}|y) \Vert_2^2\\
    &=\mathbb{E}_{p_D(y)}\mathbb{E}_{p_{D}(\boldsymbol{x}|y)}  \left\|\nabla_{\boldsymbol{x}}\log p_{\theta}(y|\boldsymbol{x})\right\| _{2}^{2} \\
    &+ 2\cdot\mathbb{E}_{p_D(y)}\mathbb{E}_{p_{D}(\boldsymbol{x}|y)} [{\rm{tr}}( \nabla_{\boldsymbol{x}}^2\log p_{\theta}(y|\boldsymbol{x}))] + {\rm const}.
 \label{ACGalignment_reform}
 \end{aligned}
 \end{equation} 
 We ignore the const in Eq.(\ref{ACGalignment_reform}) that does not depend on the model parameters and denote the DCG loss as $\mathcal{L}_{\rm{DCG}}$:
 \begin{equation}
 \begin{aligned}
    \mathop{\mathcal{L}_{\rm DCG}} &\triangleq \mathbb{E}_{p_D(y)}\mathbb{E}_{p_{D}(\boldsymbol{x}|y)}  \left\|\nabla_{\boldsymbol{x}}\log p_{\theta}(y|\boldsymbol{x})\right\| _{2}^{2} \\
    &+ 2\cdot\mathbb{E}_{p_D(y)}\mathbb{E}_{p_{D}(\boldsymbol{x}|y)} [{\rm{tr}}( \nabla_{\boldsymbol{x}}^2\log p_{\theta}(y|\boldsymbol{x}))].
 \label{ACGLoss}
 \end{aligned}
 \end{equation} 
{
Computing the Hessian trace term in Eq.\ref{ACGLoss} requires a number of {backpropagations} that is proportional to the data dimension, which is intractable for high-dimensional data. Hutchinson’s trick \cite{hutchinson1989stochastic} is a stochastic algorithm to approximate tr($\boldsymbol{A}$) for any square matrix $\boldsymbol{A}$. 
For a distribution of a random vector $\boldsymbol{v}$ such that $\mathbb{E}_{p(\boldsymbol{v})}[\boldsymbol{v}\boldsymbol{v}^\mathrm{T}]=I$, Hutchinson’s trick approximate tr($\boldsymbol{A}$) as : ${\rm{tr}}(\boldsymbol{A}) = \mathbb{E}_{p(\boldsymbol{v})}[\boldsymbol{v}^\mathrm{T}\boldsymbol{A}\boldsymbol{v}]$. Hence, using Hutchinson’s trick, we can replace ${\rm{tr}}( \nabla_{\boldsymbol{x}}^2\log p_{\theta}(y|\boldsymbol{x}))$ with $\mathbb{E}_{p(\boldsymbol{v})}[\boldsymbol{v}^\mathrm{T} \nabla_{\boldsymbol{x}}^2\log p_{\theta}(y|\boldsymbol{x})\boldsymbol{v}]$. 
Thus we can reformulate the $\mathcal{L}_{\rm{DCG}}$ as:
 \begin{equation}
 \begin{aligned}
    \mathop{\mathcal{L}_{\rm DCG}} &\triangleq \mathbb{E}_{p_D(y)}\mathbb{E}_{p_{D}(\boldsymbol{x}|y)}  \left\|\nabla_{\boldsymbol{x}}\log p_{\theta}(y|\boldsymbol{x})\right\| _{2}^{2} \\
    &+ 2\cdot\mathbb{E}_{p_D(y)}\mathbb{E}_{p_{D}(\boldsymbol{x}|y)} \mathbb{E}_{p(\boldsymbol{v})}[ \boldsymbol{v}^\mathrm{T} \nabla_{\boldsymbol{x}}^2\log p_{\theta}(y|\boldsymbol{x})\boldsymbol{v}].
 \label{ACGLoss_last}
 \end{aligned}
 \end{equation} 

{In practice, we can tune the number of samples $\boldsymbol{v}$ to trade off the performance of estimation and computational cost. With reference to the existing methods \cite{scorematching,song2019sliced}, we sample one random vector $\boldsymbol{v}$ independently for each input during the training process.} The first term in Eq.(\ref{ACGLoss_last}) can be computed by one backpropagation. The second term involves Hessian, but it is in the form of Hessian-vector products, which can be computed within $O(1)$ backpropagations. Therefore, the computation of Eq.(\ref{ACGLoss_last}) does not depend on the dimension of data and is scalable for training deep models on high-dimensional datasets. 
}

We propose fine-tuning the surrogate model by optimizing the classification loss and the {\rm DCG} loss jointly during training. The optimization objective can be formulated as:
\begin{equation}
     \mathop{minimize} \limits_{\theta}\ [\mathcal{L}(f_{\theta}(x),y)+\lambda \; \mathcal{L}_{\rm DCG}],
\label{Eq:lossACG}
\end{equation}
where $\lambda$ represents the regularization strength.

In this way, we can obtain a fine-tuned surrogate model whose gradient of log conditional density $\nabla_{\boldsymbol{\boldsymbol{x}}} \log p_\theta(y|\boldsymbol{x})$ aligns with the gradient of log {ground truth class-conditional data distribution} $\nabla_{\boldsymbol{x}} \log p_D(\boldsymbol{x}|y)$ better than the original surrogate model. Moreover, we can manipulate the distribution information of the image through the iterative adversarial attack with the fine-tuned model.

\subsection{Distribution-Relevant Attack}
{We named the attack using our distribution-relevant fine-tuned surrogate models as \textbf{D}istribution-\textbf{R}elevant \textbf{A}ttack (\textbf{DRA}).} \textbf{DRA} consists of two steps: fine-tuning the surrogate model to decrease the distance between the gradient of the model and the gradient of the ground truth data distribution, and using the fine-tuned surrogate model to generate adversarial perturbation with the guidance of the approximate gradient of the ground truth distribution. 
Alg. \ref{algorithm:DRA} details our method.

We jointly optimize the DCG loss $\mathcal{L}_{\rm{DCG}}$ and the classification loss $\mathcal{L}$ to fine-tune the surrogate model. This optimization process encourages the direction of the gradient of the surrogate model to match the direction of the gradient of the ground truth data distribution.

{Our proposed fine-tuned method aims to enable the attackers to push the image away from its original distribution using the gradient of the model. With the fine-tuned surrogate model, we can use most existing transfer attack methods to conduct attacks.
We mainly choose the widely used iterative attack method, projected gradient descent (PGD) \cite{BIM,madry2019kpgd} to generate adversarial examples. Our fine-tuning method is also compatible with other advanced transfer attacks.
}

\subsubsection{\textbf{Untargeted attack}}

The untargeted attack can be formulated as:
 \begin{equation}
 \left\{
 \begin{array}{l}   
    \boldsymbol{x_n} = \boldsymbol{x_{n-1}}+\eta\cdot sign(\nabla_{\boldsymbol{x_{n-1}}}\mathcal{L}( f_{\theta}(\boldsymbol{x_{n-1}}),y_{label}),\\
    \boldsymbol{x_n} = clip(\boldsymbol{x_n},\boldsymbol{x_0}-\epsilon,\boldsymbol{x_0}+\epsilon),
 \end{array}
 \right.
 \label{advpert_untargeted}
 \end{equation}
 where $x_n$ is the generated adversarial example after $n$ steps, and $x_0$ is the clean image. $\mathcal{L}$ is the classification loss, $\eta$ is the perturbation step size, and $y_{label}$ is the original label for the clean image.
 The $clip$ operation aims to make the perturbation bounded in the budget $\epsilon$.

 \begin{algorithm}[t]
    \caption{Distribution-Relevant Attack \textbf{DRA}: Given network $f_\theta$ parameterized by $\theta$, regularization constant $\lambda$, epochs $T$, total batches $M$, learning rate $\eta$, the classification loss $\mathcal{L}$, the DCG loss $\mathcal{L}_{{\rm DCG}}$. Adversarial perturbation $\boldsymbol{\delta}$, original image $\boldsymbol{x}$, $\ell_{\infty}$ perturbation radius $\epsilon$; step size $\alpha$; iterations $N$.}
    \begin{algorithmic}
    \STATE {$\blacktriangleright$ \textbf{Fine-tuning}}: 
    \FOR {$i = 1,2...,T$}
        \FOR {$j = 1,2...,M$} 
            \STATE Updating model parameters:
            \STATE $\theta=\theta-\eta \cdot (\nabla_{\theta}\mathcal{L}(f_{\theta}(x_j),y_j)+\lambda \cdot \nabla_{\theta}\mathcal{L}_{\rm DCG})$
        \ENDFOR
	\ENDFOR
	\STATE \textbf{return} Fine-tuned network $f_\theta$

  \STATE
  \STATE
    {$\blacktriangleright$ \textbf{Untargeted Attack}}:
        \STATE Initialize ${\boldsymbol{\delta}}=Uniform(-\epsilon,\epsilon)$.
        
        \FOR {$i = 1, 2, ..., N$}
            \STATE
            $\boldsymbol{\delta} = \boldsymbol{\delta} + \alpha\cdot sign(\nabla_{\boldsymbol{x}}\mathcal{L}(f_{\theta}({\boldsymbol{x}+\boldsymbol{\delta}}),y_{label})$,
            \STATE $\boldsymbol{\delta}=max(min(\boldsymbol{\delta},\epsilon),-\epsilon)$
    	\ENDFOR
    	\STATE \textbf{return} $\boldsymbol{\delta}$
    	
  \STATE
  \STATE
    {$\blacktriangleright$ \textbf{Targeted Attack}}:
        \STATE Initialize ${\boldsymbol{\delta}}=Uniform(-\epsilon,\epsilon)$.
        
        \FOR {$i = 1, 2, ..., N$}
            \STATE
            $\boldsymbol{\delta} = \boldsymbol{\delta} - \alpha\cdot sign(\nabla_{\boldsymbol{x}}\mathcal{L}(f_{\theta}({\boldsymbol{x}+\boldsymbol{\delta}}),y_{target})$,
            \STATE $\boldsymbol{\delta}=max(min(\boldsymbol{\delta},\epsilon),-\epsilon)$
    	\ENDFOR
    	\STATE \textbf{return} $\boldsymbol{\delta}$
    	
    \end{algorithmic}
    \label{algorithm:DRA}
\end{algorithm}

  \begin{figure}[t]
 \centering
 \includegraphics[width=0.5\textwidth]{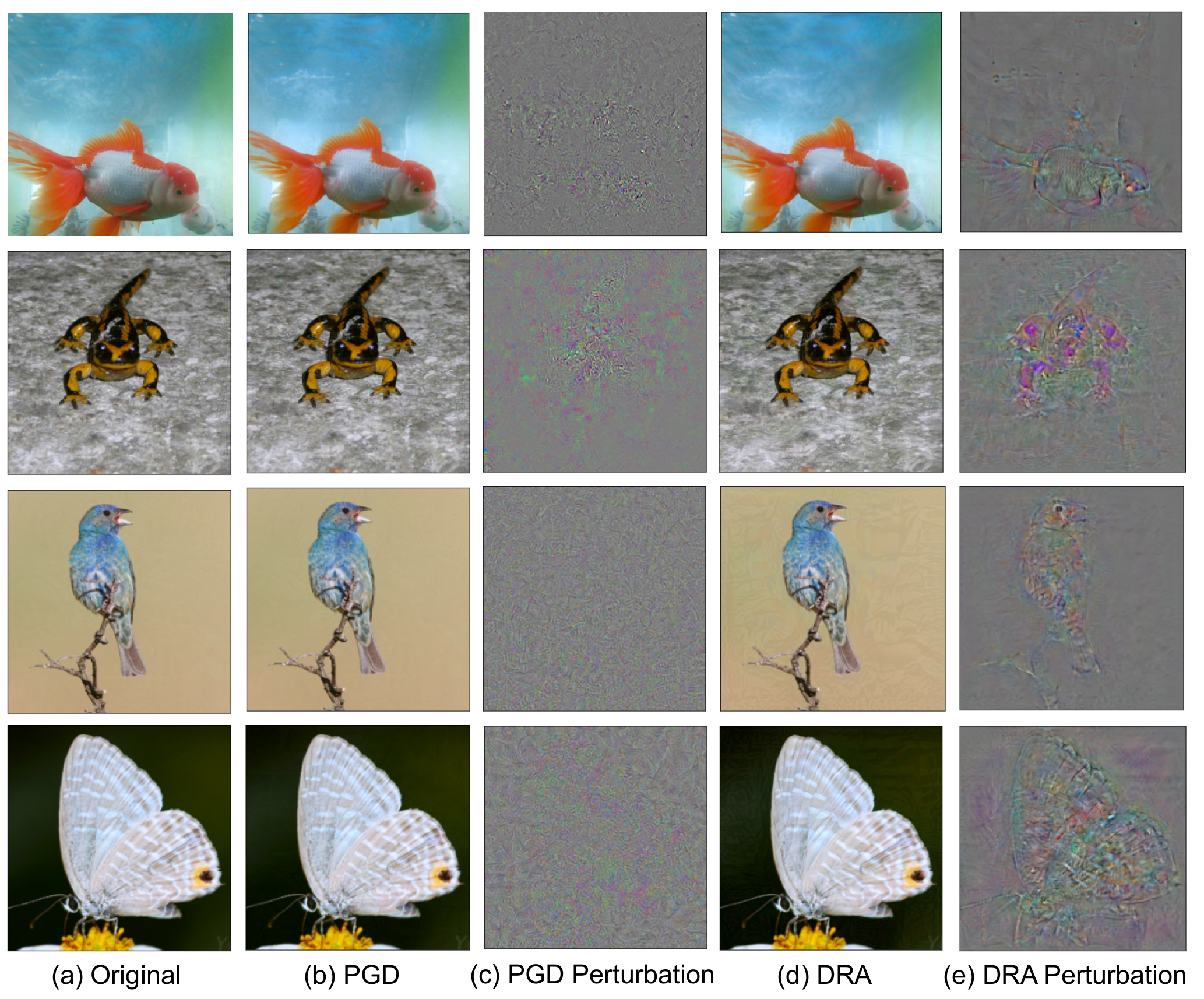}
 \caption{Comparison between the untargeted adversarial examples generated by PGD attack and our \textbf{DRA}. (a) The original images. (b) The adversarial examples generated by the PGD attack. (c) Normalizing the PGD perturbation to [0,1] for visibility. (d) The adversarial examples generated by \textbf{DRA}. (e) Normalizing the \textbf{DRA} perturbation to [0,1]. These adversarial examples are projected within a small distance (e.g., $\ell_{\infty}$ $\epsilon\leq16/255$) during inference.}
 \label{img:untargeted}
 \end{figure}

 Although DRA and other iterative methods are similar in expression when generating untargeted attacks, the direction of the \textbf{DRA} attack is better aligned with the gradient of the data distribution. 
 In other words, \textbf{DRA} moves the image out of its original distribution to generate untargeted attacks, while the other iterative methods pay attention to dragging the inputs across the decision boundary of the classifier. {As shown in Fig.\ref{img:OOD_eval}, the adversarial examples generated by our \textbf{DRA} are regarded as the out-of-distribution examples.} Fig.\ref{img:untargeted} shows the difference between the adversarial examples generated by \textbf{DRA} and PGD. 
 The adversarial perturbation of the untargeted attack generated by \textbf{DRA} concentrates on the semantic features of the image while the adversarial perturbation generated by the PGD attack seems irregular.

\begin{figure}[htbp]
\centering
\includegraphics[width=0.5\textwidth]{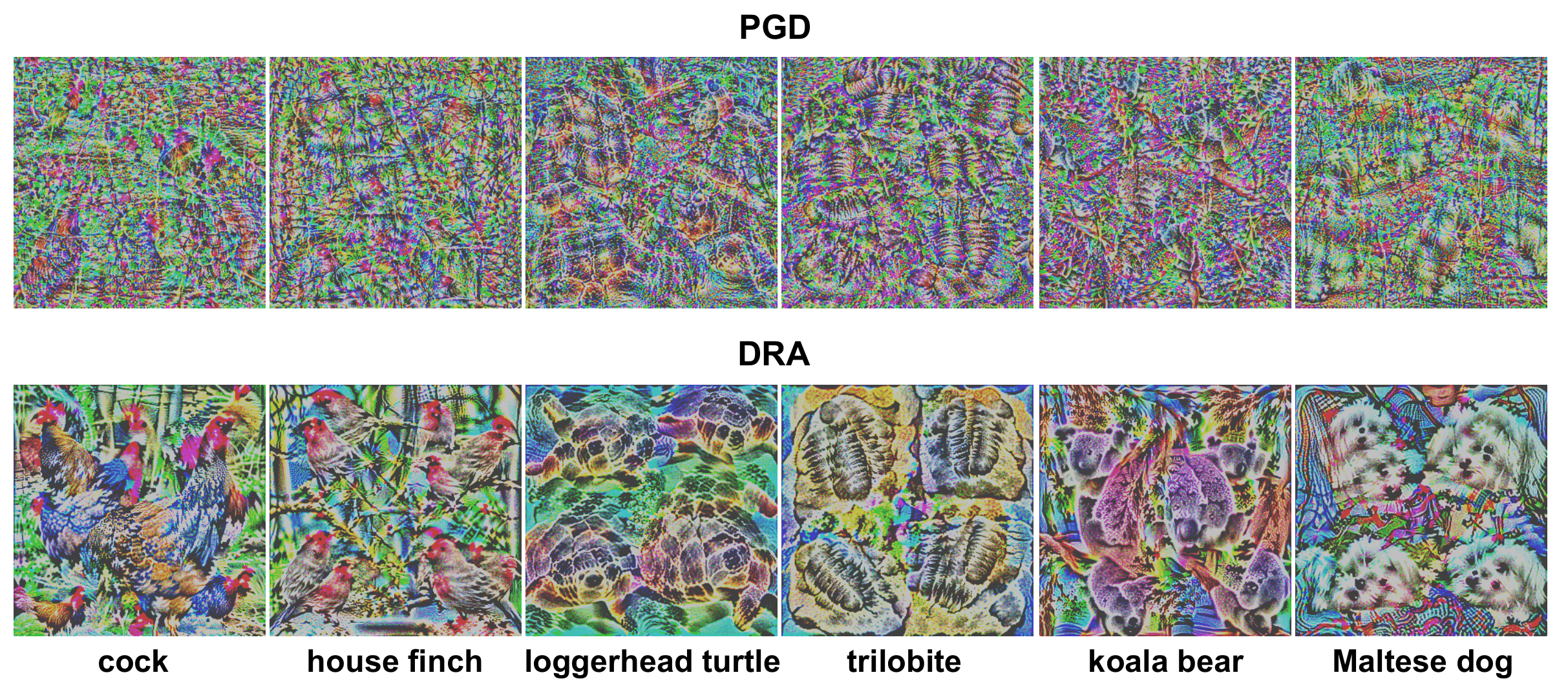}
\caption{The targeted perturbation (unbounded perturbation for visibility) generated from the mean image. The perturbation generated by our \textbf{DRA} can reflect the semantic features of the target category while the perturbation generated by the PGD attack seems more noisy.}
\label{img:targeted}
\end{figure}

\subsubsection{\textbf{Targeted attack}}
Similar to the untargeted \textbf{DRA}, the targeted version of \textbf{DRA} can be formulated as:
 \begin{equation}
 \left\{
 \begin{array}{l}   
    \boldsymbol{x_n} = \boldsymbol{x_{n-1}}-\eta\cdot sign(\nabla_{\boldsymbol{x_{n-1}}}\mathcal{L}( f_{\theta}(\boldsymbol{x_{n-1}}),y_{target}),\\
    \boldsymbol{x_n} = clip(\boldsymbol{x_n},\boldsymbol{x_0}-\epsilon,\boldsymbol{x_0}+\epsilon),
 \end{array}
 \right.
 \label{advpert_targeted}
 \end{equation}
 where attackers aim to make the DNNs misclassify the input as the target class $y_{target}$.
 We generate the targeted attack starting from the mean image\footnote{All pixel values of the mean image are set as $0.5$ out of $[0,1]$} and show the perturbation in Fig. \ref{img:targeted}. The perturbation generated by \textbf{DRA} represents recognizable features of the target distribution, while the perturbation generated by the PGD attack does not show obvious features. The targeted adversarial perturbation, which contains the sufficient features of the target category, may dominate the classification and the original features act like noise with respect to perturbations.

 Compared with the prior research on adversarial transferability by improving the attack optimization procedure, we focus on the ground truth data distribution and match the gradient of the surrogate model to the gradient of the data distribution. 
 We use the fine-tuned surrogate model to generate the iterative attack.
 In the following section, we try to validate the effectiveness of the proposed \textbf{DRA} in both untargeted and targeted attack scenarios quantitatively.


\section{Experiment}

\begin{table*}[htb]
	\centering
	\caption{Transferability against normal models: the success rates of black-box attacks (untargeted) crafted on RN50, RN152, DN121 and DN201. The maximum standard deviation of this experiment is 0.92$\%$ which is much less than our improvement. The best results are in bold.} 
	\scalebox{1.1}{ 
	\begin{tabular}{c|c|cccccccccc}  
		\hline
		 Model&Attack & VGG19 &RN152 & DN121 & DN201 & SE154 & IncV3 & IncV4 & IncRes & {ViT}\\
		\hline
		\multirow{7}*{RN50}&PGD~\cite{madry2019kpgd}&53.00$\%$&61.26$\%$&55.62$\%$&53.56$\%$&24.78$\%$&20.86$\%$&21.96$\%$&17.60$\%$& {3.82$\%$}\\
		&DI~\cite{xie2019improvingDI}&75.06$\%$&81.65$\%$&81.98$\%$&74.80$\%$&52.42$\%$&42.58$\%$&44.30$\%$&27.12$\%$&{7.12$\%$}\\
		&MI~\cite{dong2018boostingMI}&64.86$\%$&73.22$\%$&73.50$\%$&64.33$\%$&47.20$\%$&39.08$\%$&37.35$\%$&25.26$\%$&{13.34$\%$}\\
		&ILA~\cite{huang2020enhancingILA}&{83.56$\%$}& {92.46$\%$}& {88.40$\%$}& {85.24$\%$}&61.44$\%$&49.94$\%$&48.34$\%$&35.74$\%$&{11.26$\%$}\\		
		&SGM~\cite{wu2020skip}&82.72$\%$&88.40$\%$&83.56$\%$&80.34$\%$&61.30$\%$&53.72$\%$&49.83$\%$&42.86$\%$&{18.82$\%$}\\
		&IR~\cite{zhangIR}&82.46$\%$&85.24$\%$&84.35$\%$&82.10$\%$& {64.20$\%$}& {54.60$\%$}& {51.05$\%$}& {46.78$\%$} &{17.76$\%$}\\
		&\textbf{DRA+PGD}& \textbf{98.26$\%$}& \textbf{99.24$\%$}& \textbf{99.56$\%$}& \textbf{99.28$\%$}& \textbf{93.92$\%$}& \textbf{95.56$\%$}& \textbf{92.66$\%$}& \textbf{92.56$\%$}& {\textbf{58.46$\%$}}\\
		\hline
		\hline
		 Model&Attack & VGG19& RN50 & DN121 & DN201 & SE154 & IncV3 & IncV4 & IncRes &{ViT}\\
		\hline
		\multirow{7}*{RN152}&PGD~\cite{madry2019kpgd}&49.32$\%$&72.72$\%$&53.44$\%$&51.00$\%$&26.32$\%$&23.50$\%$&22.58$\%$&18.72$\%$&{5.10$\%$}\\
		&DI~\cite{xie2019improvingDI}&74.01$\%$&88.18$\%$&79.46$\%$&77.81$\%$&57.49$\%$&50.28$\%$&47.16$\%$&35.10$\%$&{10.40$\%$}\\
		&MI~\cite{dong2018boostingMI}&65.42$\%$&83.40$\%$&77.60$\%$&75.79$\%$&53.00$\%$&46.50$\%$&43.32$\%$&33.08$\%$&{15.28$\%$}\\
		&ILA~\cite{huang2020enhancingILA}&66.20$\%$&90.44$\%$&75.48$\%$&73.80$\%$&50.32$\%$&42.32$\%$&41.30$\%$&29.98$\%$&{12.26$\%$}\\		
		&SGM~\cite{wu2020skip}& {80.40$\%$}& {96.10$\%$}& {85.80$\%$}& {82.76$\%$}&61.90$\%$&53.16$\%$&49.24$\%$&43.30$\%$&{11.72$\%$}\\
		&IR~\cite{zhangIR}&73.20$\%$&92.70$\%$&83.43$\%$&80.60$\%$& {64.00$\%$}& {53.60$\%$}& {50.30$\%$}& {48.00$\%$}&{10.24$\%$}\\
		&\textbf{DRA+PGD}& \textbf{96.36$\%$}& \textbf{99.62$\%$}& \textbf{99.28$\%$}& \textbf{98.92$\%$}& \textbf{92.98$\%$}& \textbf{95.00$\%$}& \textbf{91.00$\%$}& \textbf{91.52$\%$}&{\textbf{52.08$\%$}}\\
		\hline
		\hline
		 Model&Attack & VGG19& RN50 & RN152 & DN201 & SE154 & IncV3 & IncV4 & IncRes & {ViT}\\
		\hline
		\multirow{7}*{DN121}&PGD~\cite{madry2019kpgd}&56.78$\%$&63.22$\%$&52.76$\%$&71.98$\%$&31.46$\%$&24.92$\%$&26.82$\%$&20.64$\%$&{4.62$\%$}\\
		&DI~\cite{xie2019improvingDI}&73.68$\%$&79.56$\%$&74.72$\%$&89.40$\%$&53.34$\%$&53.65$\%$&47.94$\%$&37.72$\%$&{7.54$\%$}\\
		&MI~\cite{dong2018boostingMI}&68.36$\%$&74.18$\%$&72.88$\%$&89.56$\%$&58.58$\%$&52.22$\%$&45.35$\%$&35.24$\%$&{14.84$\%$}\\
		&ILA~\cite{huang2020enhancingILA}& {87.76$\%$}& {90.38$\%$}&83.42$\%$& {95.32$\%$}&65.02$\%$&58.64$\%$&57.36$\%$&40.76$\%$&{9.60$\%$}\\
		&SGM~\cite{wu2020skip}&80.18$\%$&88.54$\%$&80.54$\%$&92.70$\%$&64.92$\%$&54.62$\%$&49.82$\%$&37.76$\%$&{12.80$\%$}\\
		&IR~\cite{zhangIR}&82.56$\%$&86.14$\%$& {85.20$\%$}&95.30$\%$& {72.20$\%$}& {62.22$\%$}& {62.10$\%$}& {56.00$\%$}&{11.58$\%$}\\
		&\textbf{DRA+PGD}& \textbf{98.32$\%$}& \textbf{99.46$\%$}& \textbf{98.78$\%$}& \textbf{99.22$\%$}& \textbf{94.80$\%$}& \textbf{95.52$\%$}& \textbf{93.60$\%$}& \textbf{92.24$\%$}&{\textbf{58.04$\%$}}\\
		\hline
		\hline
		 Model&Attack & VGG19& RN50 & RN152 & DN121 & SE154 & IncV3 & IncV4 & IncRes &{ViT}\\
		\hline
		\multirow{7}*{DN201}&PGD~\cite{madry2019kpgd}&57.76$\%$&70.68$\%$&59.08$\%$&83.06$\%$&40.60$\%$&33.80$\%$&32.46$\%$&23.80$\%$&{6.54$\%$}\\
		&DI~\cite{xie2019improvingDI}&78.11$\%$&85.34$\%$&78.18$\%$&90.20$\%$&61.75$\%$&60.04$\%$&56.15$\%$&40.56$\%$&{10.80$\%$}\\
		&MI~\cite{dong2018boostingMI}&75.09$\%$&82.46$\%$&76.39$\%$&88.18$\%$&64.38$\%$&59.62$\%$&54.85$\%$&39.40$\%$&{17.84$\%$}\\
		&ILA~\cite{huang2020enhancingILA}& {88.56$\%$}& {94.78$\%$}& {90.02$\%$}& {98.02$\%$}& {76.34$\%$}& {67.78$\%$}& {65.36$\%$}&49.50$\%$&{11.62$\%$}\\
		&SGM~\cite{wu2020skip}&82.72$\%$&91.72$\%$&86.60$\%$&96.40$\%$&72.20$\%$&62.34$\%$&56.36$\%$&45.42$\%$&{17.66$\%$}\\
		&IR~\cite{zhangIR}&76.74$\%$&90.46$\%$&85.40$\%$&95.39$\%$&73.60$\%$&59.80$\%$&63.00$\%$& {56.60$\%$}&{15.36$\%$}\\
		&\textbf{DRA+PGD}& \textbf{98.30$\%$}& \textbf{99.66$\%$}& \textbf{99.50$\%$}& \textbf{99.86$\%$}& \textbf{96.24$\%$}& \textbf{95.74$\%$}& \textbf{92.16$\%$}& \textbf{91.78$\%$}&{\textbf{57.14$\%$}}\\
		\hline
	\end{tabular}
	}
	\label{Tab:normal_models}
\end{table*}

\subsection{Implementation}
{\textbf{DRA} consists of two steps: fine-tuning the surrogate model to decrease the distance between its gradient and the gradient of the ground truth data distribution and then using the fine-tuned surrogate model to generate adversarial perturbation. 
We mainly choose the widely used PGD \cite{madry2019kpgd} attack to generate perturbation in our \textbf{DRA} in experiments. We also evaluate the compatibility of our method with the existing advanced transfer attacks in subsection D.}
All experiments in this paper are run on Tesla V100. 

{\textbf{Fine-tuning Details.} We fine-tune the pre-trained classifiers provided by PyTorch (version 1.8.0) with the SGD optimizer for 20 epochs. The learning rate is 0.001 and decays by a factor of 10 at epochs 10. The size of mini-batch is 32. We set the hyperparameter $\lambda=6$ in our method. We fine-tune the pre-trained classifiers on the training dataset of the ImageNet which is also used to train these classifiers by PyTorch to avoid data leakage problems. The training images are randomly cropped to $3\times224\times224$. The computation cost of fine-tuning the surrogate model for one epoch requires 8 hours on Tesla V100 using ResNet-50 on ImageNet.
}

\textbf{Attack Setting.} For untargeted attack scenarios, we choose the baseline attack PGD~\cite{madry2019kpgd} and $7$ state-of-the-art transfer attacks: MI attack~\cite{dong2018boostingMI}, DI attack~\cite{xie2019improvingDI}, TI attack~\cite{dong2019TI}, ILA attack~\cite{huang2020enhancingILA}, SGM attack~\cite{wu2020skip} and IR attack~\cite{zhangIR}. These methods achieve high adversarial transferability in the 
untargeted attack scenario, but their performance drops severely in the targeted attack scenario. We consider the state-of-the-art methods specially designed for the targeted attack in targeted attack scenarios, one of which is the generative method TTP~\cite{naseer2021generatingTTP}, and the other is the iteration method Simple~\cite{zhao2021_simplicity}. {We mainly evaluate these attacks on the randomly selected 5000 ImageNet \cite{Imagenet} validation images that are correctly classified by all source models. We also evaluate the performance of our method on ImageNet V2 \cite{recht2019imagenetv2} and CIFAR-10 in Sec.\ref{otherdata}.}

\textbf{Threat Model.} 
We firstly generate adversarial examples using the surrogate models and then use these adversarial examples to attack different target models.
As for the attack strength, we follow the standard attack setting for all attack methods~\cite{wu2020skip,xie2019improvingDI}. 
We set the maximum allowable adversarial perturbation as $\epsilon=16/255$ with respect to a pixel value in $[0,1]$ by default.
In untargeted attack scenarios, we set the step size $\alpha$ to $2/255$ and set the iteration steps to $N=10$. With reference to~\cite{zhao2021_simplicity}, the targeted attack needs more iterations to achieve convergence and we set the iteration steps to $N=300$. The targeted attack success rate results are averaged on 10 different target classes~\cite{naseer2021generatingTTP}.

\textbf{Target Models and Surrogate Models.} We conduct experiments on both normal target models and secured target models. For normal target models, we choose $12$ convolutional neural networks (CNNs): VGG16 (with batch normalization), VGG19 (with batch normalization)\cite{simonyan2015vgg}, ResNet-50 (RN50), ResNet-152 (RN152)~\cite{he2016deep}, DenseNet-121 (DN121), DenseNet-201 (DN201)~\cite{huang2018densenet}, 154 layers Squeeze-and-Excitation network (SE154)~\cite{hu2019senet}, Wide$\_$ResNet$\_$50$\_$2 (WRN50-2)~\cite{zagoruyko2017wideresnet}, Squeezenet1$\_$0 (SQN)\cite{iandola2016squeezenet}, shufflenet$\_$v2$\_$x1$\_$0 (SFN)\cite{ma2018shufflenet}, Inception V3 (IncV3)~\cite{szegedy2015inceptionv3}, Inception V4 (IncV4), and Inception-ResNet V2 (IncRes)~\cite{szegedy2016inceptionv4}, and we use the pretrained models in PyTorch~\cite{pytorch}. 
{We also choose the officially released ViT-B/16 \cite{dosovitskiy2020imageVIT} as the target model.}
We consider $4$ adversarially trained models, including adversarial training with $\ell_2$ perturbation and $\ell_\infty$ perturbation~\cite{robusttransfer}, and $3$ other robust training methods, including training with Styled ImageNet (SIN)~\cite{SIN}, Augmix~\cite{Augmix}, the mixture of Styled and natural ImageNet (SIN-IN), as secured methods.
We choose $5$ models as surrogate models: DNNs with skip connection (ResNet-50, ResNet-152), DNNs with dense connection (DenseNet-121, DenseNet-201) and DNNs without skip connection (VGG19).

\subsection{The Evaluation of Untargeted Attack} \label{sec:untarged}
In this section, we focus on untargeted attacks. We first conduct experiments to compare our \textbf{DRA} with other baseline methods and show the results in Tab. \ref{Tab:normal_models}.
For transfer ResNet-50 $\rightarrow$ VGG19, \textbf{DRA} achieves a success rate of 98.26$\%$, which is 45.26$\%$ and 15.54$\%$ higher than PGD and SGM, respectively. For transfer ResNet-50 $\rightarrow$ IncRes, \textbf{DRA} achieves a success rate of 92.56$\%$, which is 74.96$\%$ and 45.78$\%$ higher than PGD and IR, respectively. {Moreover, our proposed method can not only improve the adversarial transferability from the convolutional neural network (CNN) to CNN but also improve the adversarial transferability from CNN to the vision transformer. When transferring from ResNet-50 to the ViT-B/16 \cite{dosovitskiy2020imageVIT}, our \textbf{DRA} achieves the success rate of 58.46$\%$ which is 39.64$\%$ better than the previous best method SGM and 54.64$\%$ better than the baseline method PGD.} \textbf{DRA} outperforms existing methods by a large margin in all transfer scenarios. We think that generating adversarial perturbation from the perspective of data distribution is a key factor in the success of \textbf{DRA}. As shown in Fig.\ref{img:OOD_eval} and Fig.\ref{img:untargeted}, \textbf{DRA} corrupts the features of the original distribution and moves the data out of its original distribution. Thus, different models cannot give correct predictions.

\begin{figure}[htbp]
\centering
\includegraphics[width=0.495\textwidth]{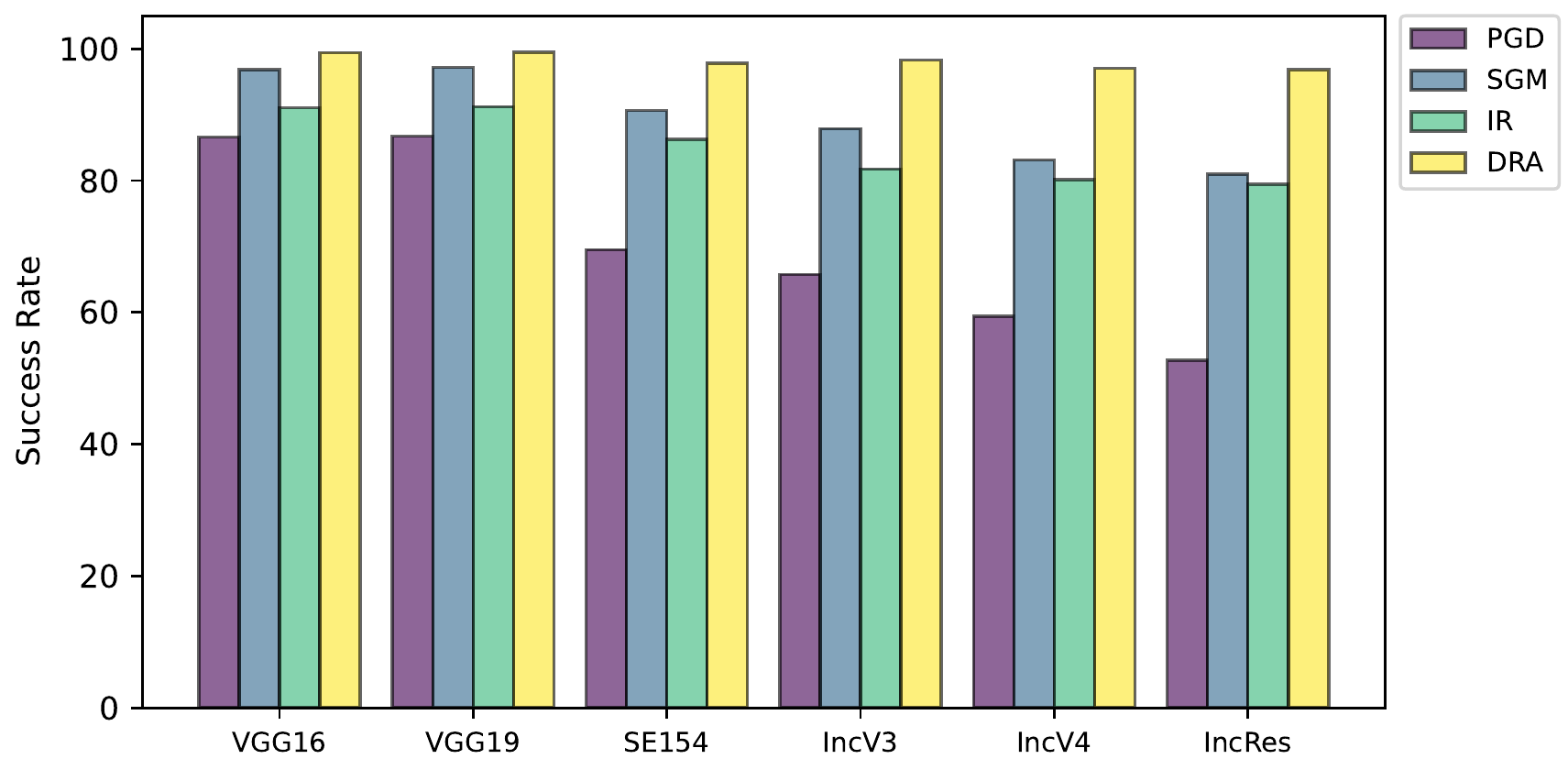}
\caption{Transferability against different models: the success rates of black-box attacks (untargeted) crafted on an ensemble of 3 models (RN50, RN152 and DN121). The horizontal axis represents different target models.}
\label{img:ensemble}
\end{figure}

As shown in Tab. \ref{Tab:normal_models}, the performance of the existing methods is unsatisfactory in some cases. Some works show that the ensemble-based strategy~\citep{liu2017delving} can improve the performance of transfer attacks~\cite{wu2020skip,zhangIR}. Fig. \ref{img:ensemble} shows the results of the ensemble strategy. We use the ensemble of ResNet-50, ResNet-152 and DenseNet-121 to generate the adversarial attack. The ensemble strategy can enhance the adversarial transferability of different methods, and our \textbf{DRA} still performs the best among the existing methods.

\begin{figure}[htbp]
\centering
\includegraphics[width=0.485\textwidth]{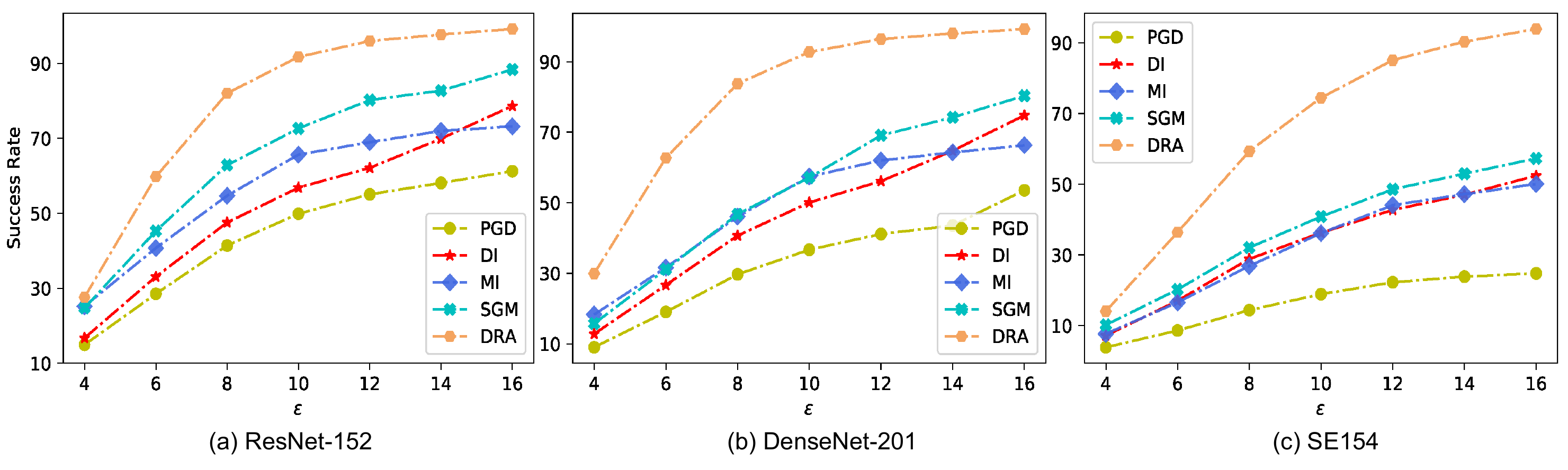}
\caption{ The attack success rate of untargeted transfer attacks on ImageNet. We use the ResNet-50 as the surrogate model and choose three different target models. The horizontal axis represents different attack strengths. Our DRA surpasses the other methods against different target models.}
\label{img1:epsilon}
\end{figure}

{The existing works always evaluate the adversarial transferability with the perturbation restricted by $\ell_\infty$ = 16/255. In Fig.\ref{img1:epsilon}, we compare our \textbf{DRA} with the existing methods with different attack strengths. \textbf{DRA} achieves the best performance with different attack strengths.}

\begin{table}[htbp]
\centering
\caption{Transferability against secured models. We generate adversarial perturbation (untargeted) by different methods. And then transfer the perturbation to ResNet-50 trained using different secured methods including Augmix~\cite{Augmix}, Stylized Imagenet\cite{SIN} and adversarial examples\cite{robusttransfer}. The best results are in bold.} 
\scalebox{0.8}{
\begin{tabular}{c|c|ccccccc}
\hline
 & Attack & SIN   & SIN-IN & Augmix & $\ell_2$=0.05 & $\ell_2$=0.1 & $\ell_\infty$=0.5 & $\ell_\infty$=1 \\
\hline
\multirow{5}{*}{{\rotatebox{90}{RN50}}}        & PGD\cite{madry2019kpgd}    & 34.44 & 84.32  & 39.64  & 24.38   & 12.74  & 3.64     & 2.88   \\
                             & MI~\cite{dong2018boostingMI}&  {56.98} & 88.74  & 64.32  &  {61.78}   &  {46.38}  &  {27.52}    &  {17.88}  \\
                             & TI~\cite{dong2019TI}& 38.46 & 80.28  & 38.74  & 26.74   & 14.12  & 4.16     & 3.12   \\
                             & SGM~\cite{wu2020skip} & 49.72 &  {96.18}  &  {72.58}  & 58.12   & 37.02  & 9.34     & 5.40    \\
                             & \textbf{DRA+PGD}    &  \textbf{98.08} &  \textbf{98.86}  &  \textbf{98.34}  &  \textbf{98.70}    &  \textbf{98.54}  &  \textbf{88.90}     &  \textbf{63.16}  \\\hline
\multirow{5}{*}{{\rotatebox{90}{DN121}}} & PGD\cite{madry2019kpgd}    & 30.74 & 45.36  & 32.46  & 20.5    & 12.20   & 4.08     & 3.06   \\
                             & MI~\cite{dong2018boostingMI}& 40.26 & 46.22  & 39.88  & 27.54   & 20.48  & 10.27    & 4.20    \\
                             & TI~\cite{dong2019TI}& 32.58 & 45.74  & 32.22  & 22.98   & 14.26  & 4.54     & 4.68   \\
                             & SGM\cite{wu2020skip} &  {43.00}    &  {58.52}  &  {53.40}   &  {44.00}      &  {34.06}  &  {13.14}    &  {8.04}   \\
                             & \textbf{DRA+PGD}    &  \textbf{97.52} &  \textbf{98.54}  &  \textbf{97.7}   &  \textbf{98.40 }   &  \textbf{98.14}  &  \textbf{89.04}    &  \textbf{66.62}  \\\hline
\multirow{5}{*}{{\rotatebox{90}{RN152}}}       & PGD\cite{madry2019kpgd}    & 29.84 & 50.76  & 35.18  & 22.16   & 12.48  & 3.92     & 3.00      \\
                             & MI~\cite{dong2018boostingMI}& 29.76 & 50.16  & 35.10   & 21.66   & 12.30   & 3.96     & 2.98   \\
                             & TI~\cite{dong2019TI}& 29.74 & 50.96  & 35.22  & 21.8    & 12.58  & 3.94     & 2.98   \\
                             & SGM\cite{wu2020skip} &  {47.62} &  {84.28}  &  {69.98}  &  {59.92}   &  {40.34}  &  {11.48}    &  {6.12}   \\
                             & \textbf{DRA+PGD}    &  \textbf{97.42} &  \textbf{98.5}   &  \textbf{97.62}  &  \textbf{97.56}   &  \textbf{77.36}  &  \textbf{79.82}    &  \textbf{50.66}  \\\hline
\multirow{5}{*}{{\rotatebox{90}{DN201}}}       & PGD\cite{madry2019kpgd}    & 34.02 & 50.76  & 39.08  & 24.56   & 14.52  & 4.46     & 3.16   \\
                             & MI~\cite{dong2018boostingMI}& 35.11 & 50.26  & 38.82  & 22.46   & 14.12  & 4.46     & 3.14   \\
                             &TI~\cite{dong2019TI}& 34.62 & 50.83  & 40.11  & 22.25   & 14.28  & 4.35     & 3.06   \\
                             & SGM\cite{wu2020skip} &  {49.14} &  {76.32}  &  {63.66}  &  {52.76}   &  {43.12}  &  {17.32}    &  {10.08}  \\
                             & \textbf{DRA+PGD}    &  \textbf{95.42} &  \textbf{98.46}  &  \textbf{97.72}  &  \textbf{96.54}   &  \textbf{58.58}  &  \textbf{65.94}    &  \textbf{33.10}  \\
                             \hline
\end{tabular}
}
\label{Tab:secured_models}
\end{table}

 In Tab. \ref{Tab:secured_models}, we provide experiments to show that the adversarial examples generated by \textbf{DRA} can also effectively penetrate the advanced secured models. Augmix~\cite{Augmix} is an augmentation-based training in order to make the model robust to natural corruptions. Training on
 Stylized ImageNet~\cite{SIN} can increase shape bias and decrease bias towards texture. Adversarial training is considered the most effective way to defend against attacks\cite{robusttransfer}. As shown in Tab. \ref{Tab:secured_models}, we generate the adversarial perturbation by different methods and then transfer the perturbation to ResNet-50 trained using different secured methods. ``$\ell_2$=0.05" represents the adversarially trained ResNet-50 using perturbation constrained in $\ell_2$ ball with radius $\epsilon=0.05$. The performance of the existing methods will severely degrade when attacking against the robust target model. For transfer ResNet-50 $\rightarrow$ robust model $\ell_\infty=1$, the attack success rate of SGM is 5.40$\%$ while our \textbf{DRA} can still achieve a success rate of 63.16$\%$. \textbf{DRA} outperforms the existing methods in attacking the advanced defense models.

\begin{table*}[htb]
	\centering
	\caption{Transferability against normal models: the success rates of black-box attacks (targeted) crafted on VGG19, RN50 and DN121. The maximum standard deviation of this experiment is 0.88$\%$ which is much less than our improvement. The best results are in bold.} 
	\scalebox{1.1}{ 
	\begin{tabular}{c|c|ccccccccc}  
		\hline
		 Model&Attack & VGG19 &DN121 & RN50 & RN152 & WRN50-2 & SQN & SFN & IncV3\\
		\hline
		\multirow{6}*{RN50}&PGD~\cite{madry2019kpgd}&1.42$\%$&2.68$\%$&93.77$\%$&2.16$\%$&2.76$\%$&2.05$\%$&1.36$\%$&0.60$\%$\\
		&DI~\cite{xie2019improvingDI}&10.64$\%$&17.54$\%$& {99.01$\%$}&13.80$\%$&13.75$\%$&1.43$\%$&1.26$\%$&4.12$\%$\\
		&Simple~\cite{zhao2021_simplicity}&70.77$\%$&57.78$\%$& \textbf{100$\%$}&59.64$\%$&68.03$\%$&7.92$\%$&6.84$\%$&15.50$\%$\\
		&CDA~\cite{naseer2019cross}&73.58$\%$&75.42$\%$&96.45$\%$&72.14$\%$&71.73$\%$&48.62$\%$&42.64$\%$&35.24$\%$\\
		&TTP~\cite{naseer2021generatingTTP}& {81.11$\%$}& {83.68$\%$}&98.13$\%$& {83.58$\%$}& {81.27$\%$}& {58.02$\%$}& {54.18$\%$}& {46.47$\%$}\\
		&\textbf{DRA+PGD}& \textbf{87.80$\%$}& \textbf{94.23$\%$}&97.03$\%$& \textbf{93.85$\%$}& \textbf{91.65$\%$}& \textbf{85.18$\%$}& \textbf{92.14$\%$}& \textbf{75.93$\%$}\\
		\hline
		\multirow{6}*{DN121}&PGD~\cite{madry2019kpgd}&1.28$\%$&97.40$\%$&1.78$\%$&1.01$\%$&1.37$\%$&2.50$\%$&1.58$\%$&0.72$\%$\\
		&DI~\cite{xie2019improvingDI}&7.31$\%$& {98.81$\%$}&9.06$\%$&5.78$\%$&6.29$\%$&1.28$\%$&1.16$\%$&1.10$\%$\\
		&Simple~\cite{zhao2021_simplicity}&50.10$\%$& \textbf{99.98$\%$}&41.74$\%$&24.90$\%$&35.09$\%$&7.57$\%$&4.13$\%$&18.53$\%$\\
		&CDA~\cite{naseer2019cross}&45.73$\%$&97.22$\%$&56.85$\%$&46.14$\%$&49.66$\%$&44.62$\%$&32.64$\%$&33.61$\%$\\		
		&TTP~\cite{naseer2021generatingTTP}& {60.71$\%$}&98.38$\%$& {71.00$\%$}& {59.12$\%$}& {59.95$\%$}& {55.42$\%$}& {36.15$\%$}& {43.14$\%$}\\
		&\textbf{DRA+PGD}& \textbf{84.83$\%$}&97.50$\%$& \textbf{92.36$\%$}& \textbf{89.84$\%$}& \textbf{88.70$\%$}& \textbf{83.63$\%$}& \textbf{85.68$\%$}& \textbf{75.48$\%$}\\
		\hline
		\multirow{6}*{VGG19}&PGD~\cite{madry2019kpgd}&95.67$\%$&0.31$\%$&0.30$\%$&0.20$\%$&0.25$\%$&0.42$\%$&0.82$\%$&0.45$\%$\\
		&DI~\cite{xie2019improvingDI}& {99.38$\%$}&3.10$\%$&2.08$\%$&1.02$\%$&1.29$\%$&1.65$\%$&1.14$\%$&0.72$\%$\\
		&Simple~\cite{zhao2021_simplicity}& \textbf{99.90$\%$}&13.79$\%$&13.55$\%$&5.16$\%$&7.50$\%$&4.45$\%$&1.39$\%$&7.01$\%$\\
		&CDA~\cite{naseer2019cross}&98.30$\%$&17.26$\%$&18.83$\%$&7.73$\%$&10.35$\%$&6.72$\%$&4.25$\%$&5.61$\%$\\		
		&TTP~\cite{naseer2021generatingTTP}&99.13$\%$& {46.58$\%$}& {48.50$\%$}& {28.55$\%$}& {33.75$\%$}& {28.32$\%$}& {5.97$\%$}& {14.79$\%$}\\
		&\textbf{DRA+PGD}&93.82$\%$& \textbf{85.65$\%$}& \textbf{84.45$\%$}& \textbf{80.48$\%$}& \textbf{75.25$\%$}& \textbf{87.25$\%$}& \textbf{82.62$\%$}& \textbf{70.58$\%$}\\
		\hline
	\end{tabular}
	}
	\label{Tab:normal_target}
\end{table*}

\subsection{The Evaluation of Targeted Attack}
 In this section, we focus on targeted attacks. Generating transferable targeted adversarial examples is much more challenging for current compared with untargeted attacks~\cite{liu2017delving,yang2021targeted}. 
 \citet{liu2017delving} show that the decision boundaries for the original class of the image of different models align well with each other, which partially explains why untargeted adversarial perturbation can transfer. However, the decision regions for the other classes of different models are very different, which causes the difficulty of obtaining high targeted adversarial transferability by attacking the surrogate model. 
 It is worth noting that different models tend to give the same predictions for images from the same distribution. As shown in Fig. \ref{img:targeted}, the targeted perturbation generated by \textbf{DRA} contains the features of the target distribution, which may dominate the prediction.
 
 Tab.\ref{Tab:normal_target} shows that our \textbf{DRA} can surpass the previous best targeted transfer attack TTP~\cite{naseer2021generatingTTP} by a large margin. For transfer ResNet-50 $\rightarrow$ VGG19, \textbf{DRA} achieves a success rate of 87.80$\%$, which is 17.03$\%$ and 6.69$\%$ higher than Simple \cite{zhao2021_simplicity} and TTP, respectively. When transferring from the surrogate model to the target model with quite different architecture, the advantages of \textbf{DRA} are more significant. For transfer ResNet-50 $\rightarrow$ IncV3, \textbf{DRA} achieves a success rate of 75.93$\%$, which is 60.43$\%$ and 29.46$\%$ higher than Simple and TTP, respectively. 
 Meanwhile, we found that Simple~\cite{zhao2021_simplicity} performs best when the surrogate and target models have the same architecture. When the architecture of the surrogate and target model is the same, Simple \cite{zhao2021_simplicity} uses the same model to generate perturbation and evaluates the attack performance, which is a white-box attack. Our \textbf{DRA} uses the modified model to generate adversarial perturbation and the target model shares the same architecture with the surrogate model but has different parameters. Thus the attack success rate of \textbf{DRA} is slightly lower than Simple in the white box attack scenario.
 However, since we focus more on the adversarial transferability between models with different architectures, we argue that this is not a conspicuous drawback of the proposed method.

 To evaluate the effectiveness of our \textbf{DRA} comprehensively, we also evaluate its targeted transferability against secured models. 
 Similar to the Sec. \ref{sec:untarged}, we consider various types of defense methods (augmented vs. stylized vs. adversarial). As shown in Tab. \ref{Tab:secured_target}, the secured models can effectively defend the adversarial examples generated by the Simple~\cite{zhao2021_simplicity} attack that is a state-of-the-art iterative targeted attack,
 while our \textbf{DRA} can still penetrate the defense model effectively.
 For example, for transfer ResNet-50 $\rightarrow$ robust model $\ell_\infty=1$, \textbf{DRA} achieves a success rate of 24.22$\%$, which is 23.98$\%$ and 22.98$\%$ higher than Simple and TTP, respectively, validating the superiority of \textbf{DRA} on transfer attacks.

\begin{table}[htbp!]
\centering
\caption{Transferability against secured models. We generate adversarial perturbation (targeted) by different methods. And then transfer the perturbation to ResNet-50 trained using different secured methods including Augmix~\cite{Augmix}, Stylized Imagenet\cite{SIN} and adversarial examples\cite{robusttransfer}. The best results are in bold.} 
\scalebox{0.78}{
\begin{tabular}{c|c|ccccccc}
\hline
 & Attack & SIN   & SIN-IN & Augmix & $\ell_2$=0.05 & $\ell_2$=0.1 & $\ell_\infty$=0.5 & $\ell_\infty$=1  \\\hline
\multirow{3}{*}{\rotatebox{90}{VGG19}} & Simple & 0.78           & 4.60& 2.64 & 0.83& 0.22           & 0.23            & 0.19  \\
            & TTP &  {47.32}  &  {61.52}   &  {77.52} &  {68.42}   &  {7.14}   &  {10.34}  &  {0.56} \\
                       & \textbf{DRA+PGD}    
                       &  \textbf{80.00}     
                       &  \textbf{89.31}
                       &  \textbf{79.73}     
                       &  \textbf{82.93}   
                       &  \textbf{41.88} 
                       &  \textbf{48.45}       
                       &  \textbf{15.34}          \\\hline
\multirow{3}{*}{\rotatebox{90}{DN121}} & Simple & 2.27           & 17.78 & 10.37  & 3.70  & 0.31 & 0.35            & 0.24  \\
          & TTP  &  {50.48}   &  {77.32}  &  {78.86}  &  {69.08}          &  {7.08}   &  {12.78}  &  {0.65}   \\
                       & \textbf{DRA+PGD}    &  \textbf{78.83} &  \textbf{92.92} &  \textbf{86.74} &  \textbf{90.22} &  \textbf{52.52} &  \textbf{58.43} &  \textbf{18.01} \\\hline
\multirow{3}{*}{\rotatebox{90}{RN50}}  & Simple & 4.29           & 67.11 & 23.08& 7.08   & 0.34  & 0.53            & 0.24  \\
          & TTP &  {57.75}  &  {92.96} &  {88.79} &  {74.95} &  {7.62} &  {14.23} &  {1.24} \\
                       & \textbf{DRA+PGD}    &  \textbf{88.62} &  \textbf{97.41} &  \textbf{92.21} &  \textbf{94.92} &  \textbf{59.42} &  \textbf{67.00}     &  \textbf{24.22}\\\hline
\end{tabular}
}
\label{Tab:secured_target}
\end{table}

\begin{table*}[htbp]
\centering
\caption{The attack success rates of untargeted black-box attacks on ImageNet crafted on RN50 and DN121. The maximum standard deviation of this experiment is 0.96$\%$ which is much less than our improvement. The best results are in bold.} 
\scalebox{1.15}{ 
\begin{tabular}{cccccccccc}
\hline
Source & Attack & Vgg19 & RN152 & DN121 & DN201 & SE154 & IncV3 & IncV4 & IncRes \\
\hline
\multirow{12}{*}{ResNet50} & PGD \cite{madry2019kpgd}& 53.00 & 61.26 & 55.62 & 53.56 & 24.78 & 20.86 & 21.96 & 17.60 \\
 & PGD+DRA & 98.26 & \textbf{99.24} & \textbf{99.56} & \textbf{99.28} & 93.92 & 95.56 & 92.66 & 92.56 \\
 \cline{2-10}
 & MI \cite{dong2018boostingMI}& 64.86 & 73.22 & 73.50 & 64.33 & 47.20 & 39.08 & 37.35 & 25.26 \\
 & MI+DRA & 96.36 & 97.88 & 98.68 & 92.28 & 92.16 & 94.06 & 91.94 & 91.78 \\
  \cline{2-10}
 & NI \cite{lin2020nesterov_NI-FGSM} & 79.20 & 84.20 & 81.56 & 78.20 & 55.62 & 47.96 & 50.14 & 39.56 \\
 & NI+DRA & 96.52 & 98.34 & 99.00 & 98.44 & 91.76 & 93.30 & 90.04 & 90.14 \\
  \cline{2-10}
 & SGM \cite{wu2020skip}& 82.87 & 88.40 & 83.56 & 80.34 & 61.30 & 53.72 & 49.83 & 42.86 \\
 & SGM+DRA & 97.52 & 98.58 & 99.00 & 98.40 & 93.04 & 94.56 & 91.42 & 90.42 \\
  \cline{2-10}
 & SI \cite{lin2020nesterov_NI-FGSM} & 75.22 & 86.46 & 83.56 & 79.30 & 46.46 & 42.68 & 41.68 & 31.50 \\
 & SI+DRA & 96.80 & 98.88 & 99.24 & 98.82 & 92.72 & 95.62 & 92.94 & 93.82 \\
  \cline{2-10}
 & DI \cite{xie2019improvingDI}& 75.06 & 81.65 & 81.98 & 74.80 & 52.42 & 42.58 & 44.30 & 27.12 \\
 & DI+DRA & \textbf{98.88} & 98.94 & 99.34 & 99.06 & \textbf{96.28} & \textbf{96.80} & \textbf{95.02} & \textbf{94.02} \\
 \hline
 \hline
Source & Attack & Vgg19 & RN50 & RN152 & DN201 & SE154 & IncV3 & IncV4 & IncRes \\
\hline

\multirow{12}{*}{DenseNet121} & PGD \cite{madry2019kpgd} & 56.78 & 58.22 & 43.76 & 65.98 & 23.46 & 21.92 & 23.82 & 17.64 \\
 & PGD+DRA & 98.32 & \textbf{99.46} & \textbf{98.78} & 99.22 & 94.80 & 95.52 & 93.60 & 92.24 \\
  \cline{2-10}
 & MI \cite{dong2018boostingMI}& 68.36 & 74.18 & 67.88 & 79.56 & 55.58 & 49.22 & 42.35 & 32.24 \\
 & MI+DRA & 96.60 & 98.44 & 97.28 & 98.38 & 92.30 & 93.62 & 91.88 & 89.90 \\
  \cline{2-10}
 & NI \cite{lin2020nesterov_NI-FGSM} & 83.72 & 86.08 & 74.98 & 89.84 & 62.36 & 54.66 & 56.90 & 43.78 \\
 & NI+DRA & 96.98 & 98.88 & 97.96 & 98.76 & 92.62 & 92.78 & 90.40 & 88.18 \\
  \cline{2-10}
 & SGM \cite{wu2020skip}& 80.18 & 88.54 & 80.54 & 92.70 & 64.92 & 54.62 & 49.82 & 37.76 \\
 & SGM+DRA & 92.16 & 94.76 & 90.98 & 93.04 & 82.60 & 79.58 & 76.18 & 70.22 \\
  \cline{2-10}
 & SI \cite{lin2020nesterov_NI-FGSM} & 76.58 & 80.90 & 69.08 & 88.64 & 48.68 & 46.08 & 45.24 & 33.52 \\
 & SI+DRA & 97.64 & 99.28 & 98.66 & \textbf{99.34} & 94.74 & 95.90 & 94.06 & \textbf{93.68} \\
  \cline{2-10}
 & DI \cite{xie2019improvingDI}& 73.68 & 79.56 & 71.72 & 80.40 & 53.34 & 49.65 & 43.94 & 31.72 \\
 & DI+DRA & \textbf{98.94} & 99.34 & 98.42 & 99.18 & \textbf{95.90} & \textbf{95.76} & \textbf{94.78} & 92.16\\
 \hline
\end{tabular}
}
\label{tab2:combine}
\end{table*}

{\subsection{The Compatibility of DRA with Other Attacks}
Various techniques have been proposed to improve the transferability of adversarial attacks, such as advanced gradient calculations \cite{dong2018boostingMI,lin2020nesterov_NI-FGSM,wu2020skip}, and input transformations \cite{xie2019improvingDI,lin2020nesterov_NI-FGSM}. In the previous subsections, we mainly show that our \textbf{DRA} method can significantly improve the performance of the widely used baseline attack PGD\cite{madry2019kpgd}. Here, we delve into the compatibility of our \textbf{DRA} method with other attack methods. In Tab.\ref{tab2:combine}, we compare the adversarial transferability of the original adversarial attacks and the DRA version of these attacks. "PGD" means the PGD attack generated with the normal pre-trained surrogate models and "PGD+DRA" means the PGD attack generated by our \textbf{DRA} fine-tuned surrogate models. Our \textbf{DRA} fine-tuning method can significantly improve the performance of different baseline attacks. Moreover, incorporating \textbf{DRA} with the input transformations based attacks (SI, DI) achieves better performance than the advanced gradient based attacks (MI, NI, SGM). We think the advanced gradient based attacks somewhat change the gradient of the model which may increase the DCG in our fine-tuned models.
}

\begin{figure}[hbt]
 \includegraphics[width=0.495\textwidth]{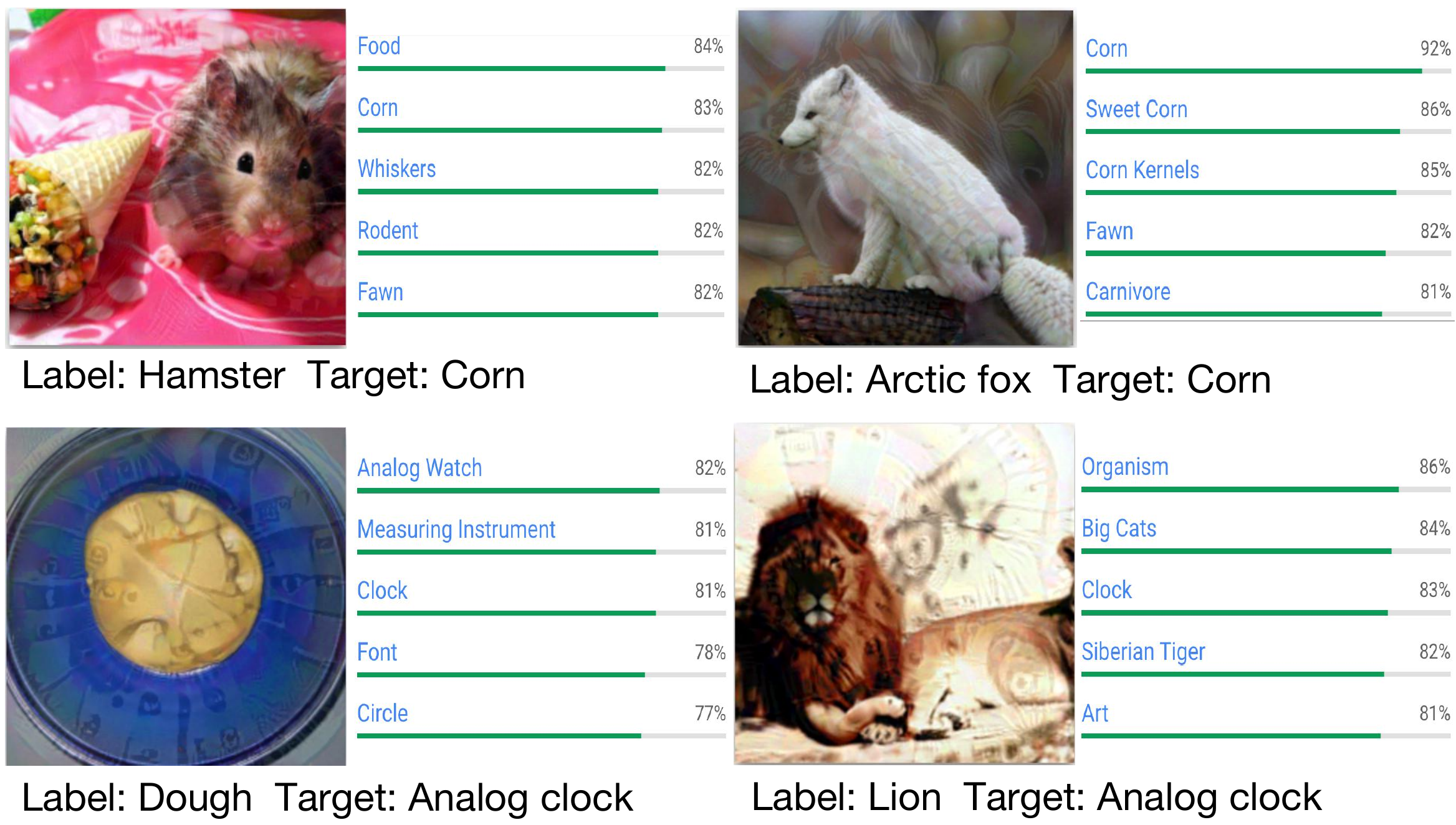}
 \centering
 \caption{We show that the adversarial perturbation generated by our \textbf{DRA} is imperceptible to human observers but can successfully fool the Google Cloud Vision with the target class.}
 \label{img:googleattack}
 \end{figure}

{\subsection{Transfer-based Attack on Google Cloud Vision}
In this section, we evaluate our DRA attack in the more challenging case, attacking a real-world computer vision system (the Google Cloud Vision API). Most existing works fool real-world computer vision systems with query-based attacks, which require a large number of queries \cite{faceattack_query,brendel2018decision_query,ilyas2018black_query}. In contrast, we apply the transfer-based attack to fool the Google Cloud Vision API. Specifically, we use the ResNet-50 as the surrogate model to generate the adversarial examples and then use the Google Cloud Vision API to predict these examples. The API predicts a list of semantic labels along with confidence scores. We measure both the targeted and untargeted transferability. For the untargeted attack, we measure whether or not the ground-truth class appeared in the returned list. For the targeted attack, we measure whether or not the target class appeared in the returned list. Since the predicted label space of Google Cloud Vision API does not precisely correspond to the 1000 ImageNet classes, we treated semantically similar classes as the same, following the setting in \cite{zhao2021_simplicity}. We take the evaluation on randomly selected 500 images that originally yield correct predictions. Fig.\ref{img:googleattack} shows the targeted adversarial examples generated by our method and the top-5 predictions made by the Google Cloud Vision API, the hamster and the arctic fox are misclassified as corn by the Google Cloud Vision API. We compare our method with the original PGD and the previous best iterative transfer attack method Simple \cite{zhao2021_simplicity} in Tab.\ref{tab:googleattack}. 
In particular, \textbf{DRA} achieves the best attack performance compared with previous attack methods. This demonstrates the high practicality of our method which can even attack real-world computer vision systems with a high success rate, e.g., Google Cloud Vision API.}

 \begin{table}[hbt]
 \centering
\caption{Untargeted and targeted transfer attack success rates ($\%$) of different attacks on Google Cloud Vision.}
\scalebox{1.25}{ 
\begin{tabular}{cccc}
\hline
 & PGD \cite{madry2019kpgd} & 
 Simple\cite{zhao2021_simplicity} & DRA \\
\hline
Untargeted & 50.6 & 51.8 & 86.6 \\
Targeted & 8.2 & 19.4 & 48.6 \\
\hline
\end{tabular}
} \label{tab:googleattack}
\end{table}

{\subsection{Evaluation on the Other Datasets}\label{otherdata}

\subsubsection{The Evaluation on ImageNet-V2}

In this subsection, we evaluate our method on the ImageNet-V2 with 10000 images \cite{recht2019imagenetv2} which are independent of existing pre-trained models. As shown in Tab.\ref{img:ImageNetV2}, our \textbf{DRA} method surpasses the other methods by a large margin. As for transfer from ResNet-50 to Inception-V3, \textbf{DRA} can improve the performance of the baseline method PGD from $59.47\%$ to $99.24\%$.
}

\begin{table*}[htbp]
\centering
\caption{The attack success rates of untargeted black-box attacks on ImageNet V2 crafted on RN50 and DN121. The maximum standard deviation of this experiment is 0.84$\%$ which is much less than our improvement. The best results are in bold.} 
\scalebox{1.15}{ 
\begin{tabular}{cccccccccc}
\hline
Source & Attack & VGG19 & RN152 & DN121 & DN201 & SE154 & IncV3 & IncV4 & IncRes \\
\hline
\multirow{5}{*}{ResNet50} &PGD \cite{madry2019kpgd} & 83.34 & 86.20 & 84.04 & 80.29 & 65.51 & 59.47 & 59.43 & 52.45 \\
 & DI \cite{xie2019improvingDI}& 95.44 & 89.95 & 90.72 & 87.91 & 79.29 & 74.16 & 74.47 & 62.47 \\
 & MI \cite{dong2018boostingMI}& 92.67 & 94.34 & 94.30 & 93.18 & 82.87 & 75.98 & 74.34 & 67.25 \\
 & SGM \cite{wu2020skip}& 95.34 & 95.52 & 93.84 & 91.55 & 82.99 & 75.56 & 73.14 & 62.88 \\
 & \textbf{DRA+PGD} & \textbf{99.39} & \textbf{99.66} & \textbf{99.81} & \textbf{99.76} & \textbf{98.11} & \textbf{99.24} & \textbf{98.30} & \textbf{98.14} \\
 \hline
 \hline
Source & Attack & VGG19 & RN50 & RN152 & DN201 & SE154 & IncV3 & IncV4 & IncRes \\
\hline
\multirow{5}{*}{DenseNet121} & PGD \cite{madry2019kpgd} & 86.23 & 87.13 & 80.21 & 90.07 & 68.57 & 63.02 & 63.58 & 54.71 \\
 & DI \cite{xie2019improvingDI}& 95.35 & 91.41 & 84.41 & 91.89 & 80.16 & 73.85 & 76.17 & 63.96 \\
 & MI \cite{dong2018boostingMI}& 93.77 & 93.99 & 89.71 & 96.41 & 83.92 & 77.22 & 76.38 & 68.93 \\
 & SGM \cite{wu2020skip}& 94.91 & 95.89 & 91.51 & 96.78 & 84.75 & 76.58 & 75.67 & 64.97 \\
 & \textbf{DRA+PGD} & \textbf{99.41} & \textbf{99.71} & \textbf{99.62} & \textbf{99.82} & \textbf{98.04} & \textbf{98.76} & \textbf{98.09} & \textbf{97.52} \\
 \hline
\end{tabular}
}\label{img:ImageNetV2}
\vspace{0.2cm}
\end{table*}

\subsubsection{The Evaluation on CIFAR-10}

\begin{figure}[t]
\centering
\includegraphics[width=0.495\textwidth]{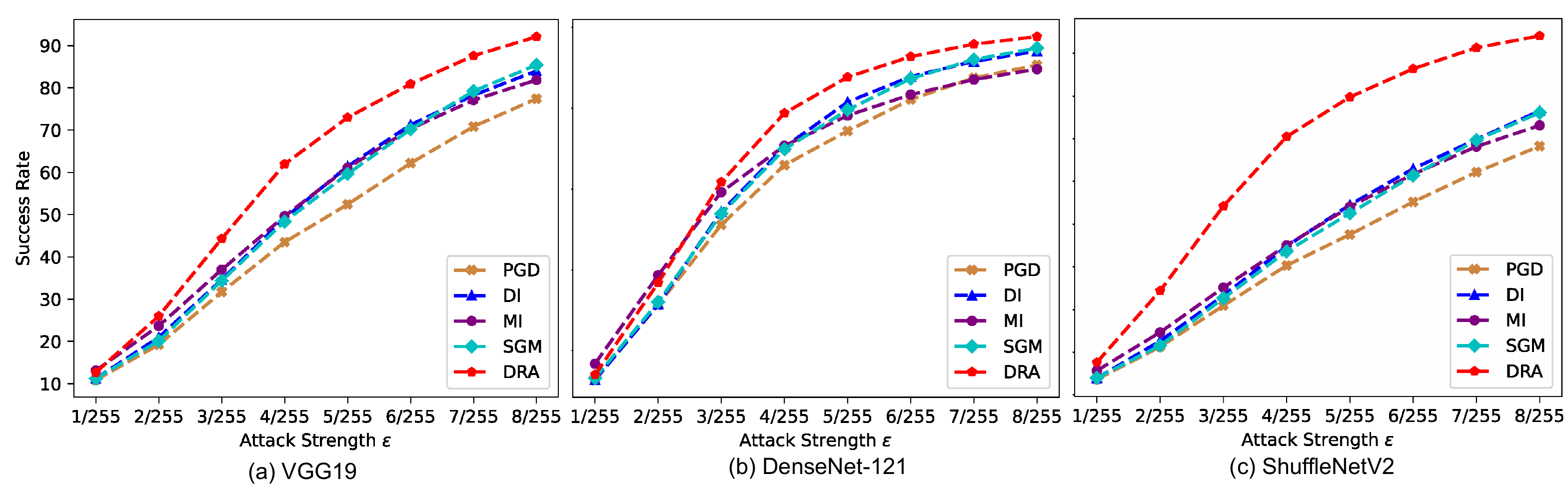}
\caption{ The attack success rate of transfer attacks on CIFAR-10. We illustrate the attack success rate for different methods on CIFAR-10. We use the ResNet-18 as the surrogate model and choose three different target models. The horizontal axis represents different attack strengths. Our DRA surpass the other methods against different target models.}
\label{img:cifar10}
\end{figure}

Following the existing works \cite{wu2020skip,zhangIR,zhao2021_simplicity}, we focus on addressing the transferability on the ImageNet dataset in the previous subsection. 
In this subsection, we conduct experiments on another dataset (CIFAR-10) to verify the effectiveness of the \textbf{DRA} algorithm further.
We use the ResNet-18 as the surrogate model and choose three target models with different architecture (VGG19, DenseNet-121 and ShuffleNetV2). We set the step size $\alpha$ to $2/255$, and set the iteration steps to $N=10$. As shown in Fig. \ref{img:cifar10}, our \textbf{DRA} shows better transferability than existing transfer attacks. For transfer ResNet-18 $\rightarrow$ VGG19 with attack strength $\epsilon=4/255$, \textbf{DRA} can achieve a success rate of 61.97$\%$ which is 13.62$\%$ higher than SGM \cite{wu2020skip} and 18.48$\%$ higher than PGD \cite{madry2019kpgd}. So far, we have evaluated the effectiveness of our \textbf{DRA} on the large-scale datasets (ImageNet) and small-scale datasets (CIFAR-10).

\section{Discussion}\label{AdditionalAnalysis}

{\subsection{Understanding the superiority of DRA} The superiority of \textbf{DRA} can be understood from two aspects. First, \textbf{DRA} reduces the dependency on the surrogate model. The existing transfer attacks usually regard the over-fitting on the surrogate models as the hindering factor of adversarial transferability and devote to alleviating the over-fitting by improving the optimization algorithm. Our proposed method seeks the commonality among different models from the data distribution perspective for that the ground truth data distribution is model-independent. Our method alleviates this over-fitting by aligning the gradient of the model with the gradient of the ground-truth data distribution. In this way, \textbf{DRA} can effectively reduce the dependence on the surrogate model and generate high transferable adversarial examples. Fig.\ref{img:pcc} illustrates the frequency histograms of Pearson Correlation Coefficient (PCC) \cite{benesty2009pearson} between the adversarial perturbations generated through different models with the same input. The correlation of the adversarial perturbations generated through different models using the PGD \cite{madry2019kpgd} attack is around zero, which confirms that the perturbations generated by the original PGD is specific to different surrogate models. The adversarial perturbations generated by different models using our \textbf{DRA} show a stronger correlation than other existing methods, which means that our \textbf{DRA} can effectively reduce the dependency of the perturbation on the surrogate model.

\begin{figure}[htbp]
\centering
\includegraphics[width=0.495\textwidth]{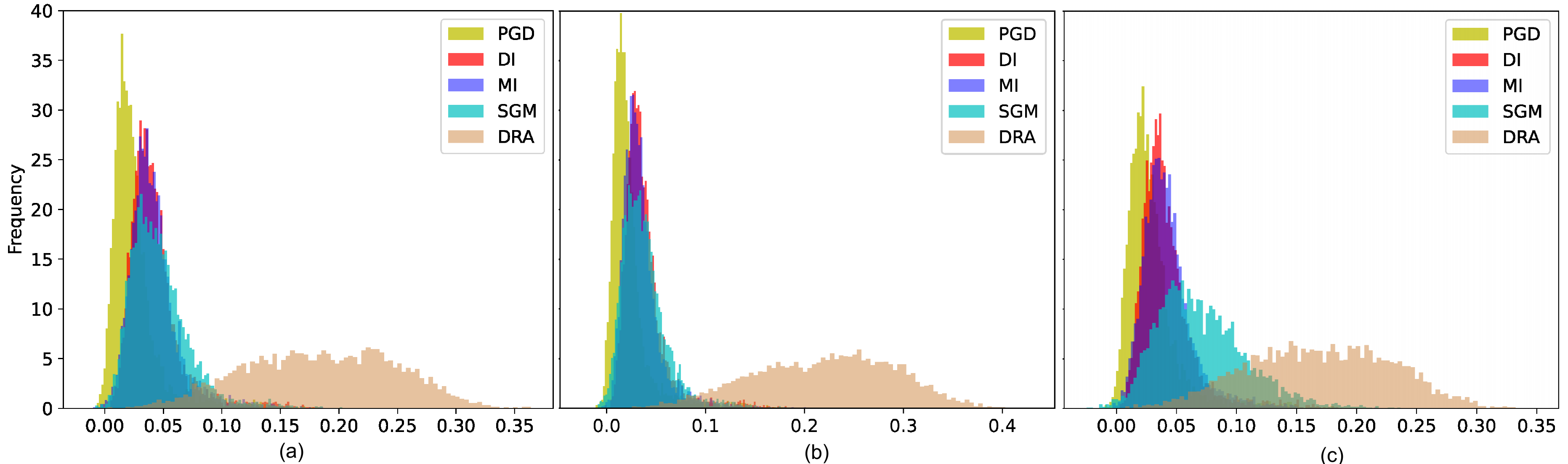}
\caption{The frequency histogram of the Pearson correlation coefficient (PCC) between adversarial perturbations of the same image generated by the surrogate models ResNet-50 and ResNet-152 (a), ResNet-50 and DenseNet-121 (b), and DenseNet-121 and DenseNet-201 (c). Higher PCC indicates a greater positive correlation. The perturbations generated by our \textbf{DRA} method are closer to each other with different surrogate models than perturbations generated by other methods.}
\label{img:pcc}
\vspace{5pt}
\end{figure}

Second, our proposed method intrinsically changes the distribution of input images, leading to more transferable adversarial images.
To be specific, our untargeted attack moves the image out of its original distribution, making it difficult for classifiers to classify the image correctly. Fig.\ref{img:OOD_eval} shows that the untargeted adversarial examples generated by our \textbf{DRA} are regarded as the out-of-distribution examples by the OOD detection method \cite{liu2020energy}. Out-of-distribution examples can mislead the deep models and cause safety concerns \cite{liu2020energy,huang2021importance}.
Our targeted attack can effectively iteratively imprint the features of the target distribution on the image, leading different classifiers to misclassify the image as the target class. For example, Fig.\ref{img:corn} shows that the targeted adversarial perturbation generated by our method contains recognizable features of the target distribution that dominates the classification. However, the adversarial example generated by the normal PGD \cite{madry2019kpgd} does not contain the semantic features of the target class.

}

{\subsection{Compared with Other Fine-tuning Methods}
Our fine-tuning method aims to match the direction of the adversarial attack to the gradient of the ground truth data distribution, which is different from the other fine-tuning methods that aim to improve the generalization of models. In Tab.\ref{tab1:augmix_mealv2}, we compare the test accuracy and the adversarial transferability of different ResNet-50 models. ``AugMix" means the ResNet-50 fine-tuned with the data processing technique AugMix \cite{Augmix}, ``AutoAug" means the ResNet-50 fine-tuned with fast AutoAugment \cite{lim2019fastAutoAugment} and ``MEALV2" means the ResNet-50 fine-tuned with the knowledge distillation method MEALV2 \cite{shen2020mealv2} that achieves 80$\%$+ Top-1 accuracy on ResNet-50. Our method reduces the test accuracy of the surrogate model but greatly enhances the adversarial transferability. This experiment shows that the generalization (test accuracy) of the model may not the key factor for adversarial transferability.
}

\begin{table}[htbp]
\centering
\caption{The Top-1 test accuracy and the success rates of untargeted PGD attacks for different ResNet-50 models. The best results are in bold.}
\begin{tabular}{c|c|ccccc}
\hline
Model & Acc & VGG19 & RN152 & DN121 & SE154 & IncV3 \\\hline
Original & 76.13 & 53.00 & 61.26 & 55.62 & 24.78 & 20.86 \\\hline
AugMix & 77.53 & 55.90 & 56.46 & 54.18 & 34.10 & 27.72 \\\hline
AutoAug & 77.60 & 50.18 & 45.76 & 39.14 & 19.12 & 16.48 \\\hline
MEALV2 & \textbf{80.67} & 42.40 & 35.32 & 39.16 & 25.56 & 19.38 \\\hline
DRA & 61.06 & \textbf{98.26} & \textbf{99.24} & \textbf{99.56} & \textbf{93.92} & \textbf{95.56}\\\hline
\end{tabular}\label{tab1:augmix_mealv2}
\vspace{5pt}
\end{table}

\subsection{The influence of the hyperparameter}
\begin{figure}[htbp]
\centering
\includegraphics[width=0.49\textwidth]{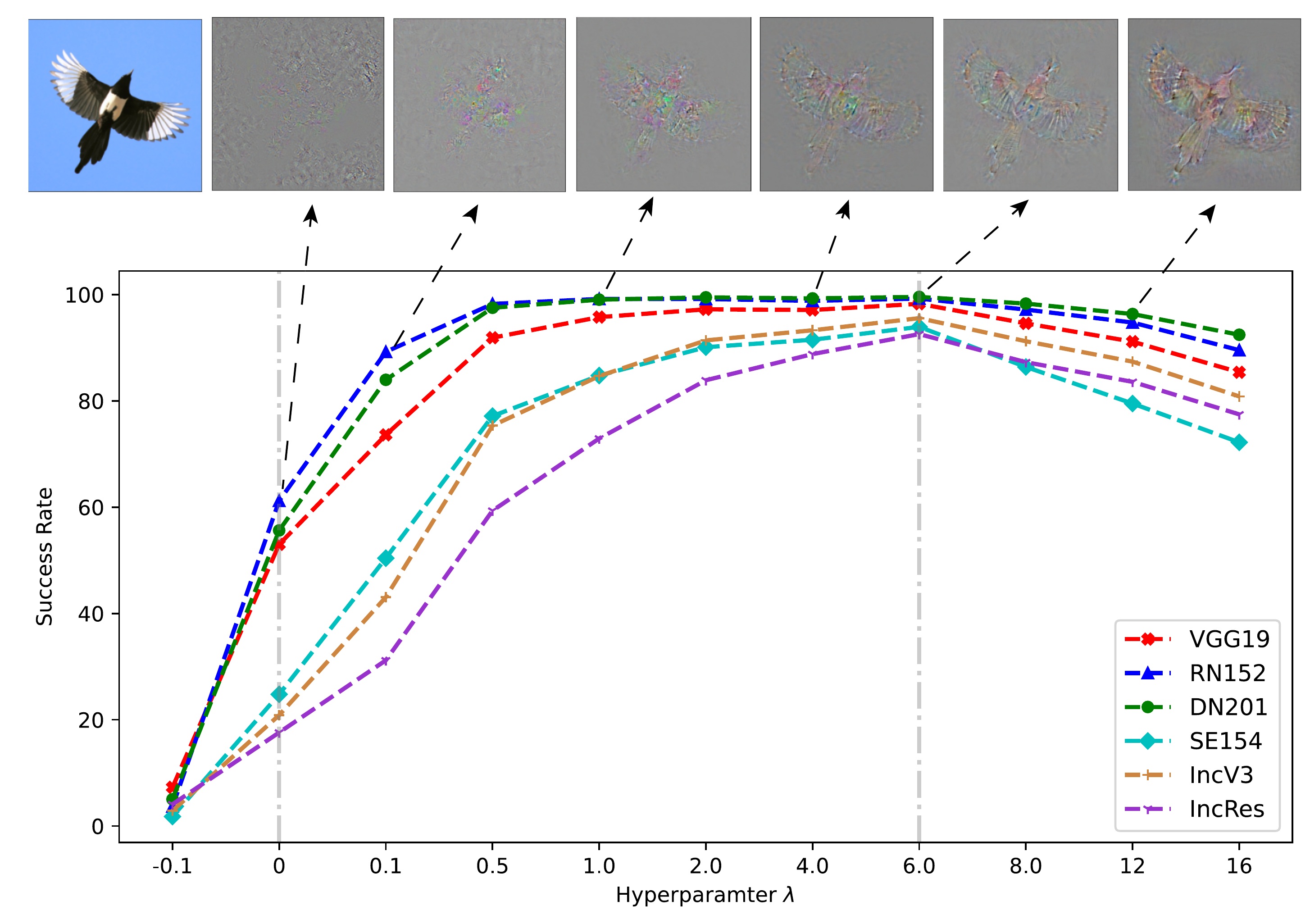}
\caption{We illustrate the attack success rate (untargeted) when generating adversarial perturbation by different \textbf{DRA} fine-tuned models. ``$\lambda$=1.0" represents the model fine-tuned with hyperparameter $\lambda$=1. We also present a seagull's adversarial perturbation generated by different \textbf{DRA} fine-tuned models.}
\label{img:lambda}
\end{figure}

In this subsection, we show that decreasing the distance between the gradient of the surrogate model and the gradient of the ground truth data distribution enhances adversarial transferability, whereas increasing this distance has the opposite effect.
As shown in Eq. \ref{Eq:lossACG} and our Alg. \ref{algorithm:DRA}, our proposed method \textbf{DRA} has one hyperparameter $\lambda$. We optimize the classification loss and the DCG loss jointly during fine-tuning and adjust the strength of DCG loss through the hyperparameter $\lambda$. Fig. \ref{img:lambda} shows the influence of $\lambda$. If $\lambda$ is less than 0, the distance between the input-gradients of the surrogate model and the gradients of the ground truth data distribution is increased, and the adversarial transferability drops dramatically. If $\lambda$ is greater than 0, the distance between the input-gradients of the surrogate model and the gradients of the ground truth data distribution is decreased, and the adversarial transferability is improved. Meanwhile, the relationship between the adversarial transferability and $\lambda$ is not monotonic. Too large $\lambda$ may restrict the model's capability of learning classification-related features. 

\subsection{t-SNE visualizations}
\begin{figure}[htbp]
\centering
\includegraphics[width=0.495\textwidth]{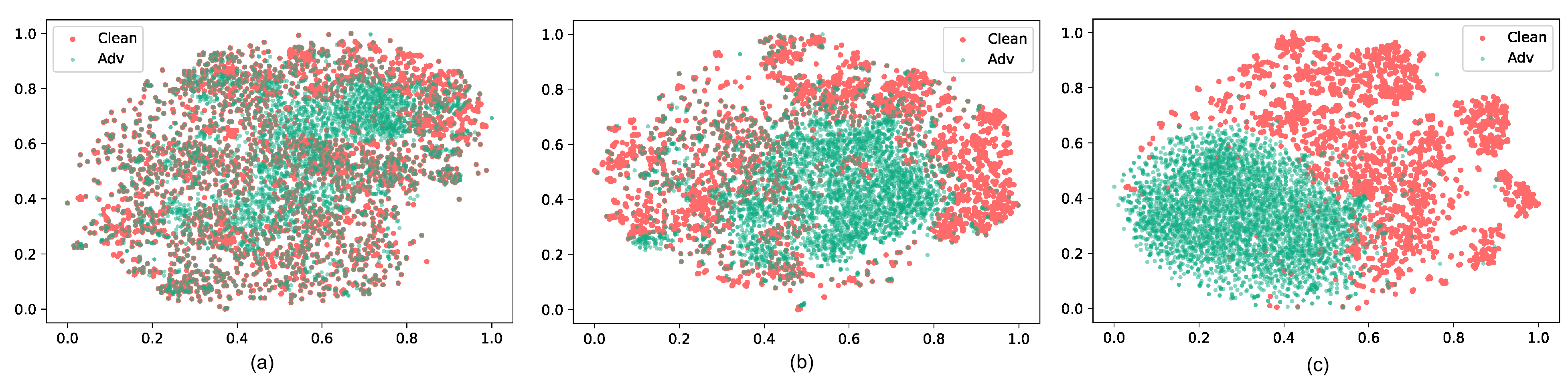}
\caption{t-SNE visualizations. We illustrate the t-SNE plots for clean (red) images and their adversarial (green) examples. (a) PGD. (b) SGM. (c) \textbf{DRA} (Ours).}
\label{img:tsne}
\end{figure}

To find out whether the generated adversarial examples have attacked the target model in the desired way. We visualize the feature embeddings of both the clean images (red) and their corresponding adversarial examples (green) using t-SNE~\cite{tsne}.
To be specific, we use the ResNet-50 (surrogate model) to generate untargeted adversarial examples and use the DenseNet-121 (target model) to obtain the final latent representation.
Further distance indicates better performance for the untargeted attack.
As shown in Fig.~\ref{img:tsne}, though the final latent representations of the adversarial examples generated by SGM show more difference with the clean images than PGD, our method yields features that clearly separate the clean and adversarial images, compared with PGD \cite{madry2019kpgd} and SGM \cite{wu2020skip}. This further validates the superiority of our proposed \textbf{DRA}.

\section{Conclusion}
{In this paper, we propose a novel understanding of adversarial transferability from the data distribution perspective.
We find that moving the image out of its original distribution can enhance the untargeted adversarial transferability and dragging the image toward the target distribution contributes to high targeted adversarial transferability.
We propose a method named \textbf{D}istribution-\textbf{R}elevant \textbf{A}ttack (\textbf{DRA}), which apply a fine-tuned surrogate model to generate more transferable adversarial images with data-distribution relevant information. Technically, we propose to fine-tune the surrogate model with a gradient matching method to match the gradient of the model and the gradient of the data distribution, which enables us to push the image away from its original distribution with the gradient of the model.
Extensive experiments demonstrate that the proposed \textbf{DRA} establishes state-of-the-art performance in both untargeted and targeted attack scenarios. Moreover, \textbf{DRA} can also effectively fool the real-world computer vision system (the Google Cloud Vision API).
Our finding not only motivates researchers to rethink the adversarial transferability from a data distribution perspective but also provides a strong counterpart for future research on adversarial defense.
}

\section*{Acknowledgments}
This work was supported in part by the Fundamental Research Funds for the Central Universities, and by Alibaba Group through Alibaba Research Intern Program, and by the Natural Science Foundation of China under Grant 62102386.



{\appendices

\section*{Proof of the intergration by parts formula} 
We assume that $g(\boldsymbol{x})$ and $f(\boldsymbol{x})$ are differentiable in Eq. (\ref{integration_part}). We can get the following equation:

\begin{footnotesize}
\begin{equation}
\begin{aligned}
 \nabla_{{x_i}}[g(\boldsymbol{x})\nabla_{{x_i}}f(\boldsymbol{x})]  &=  \nabla_{{x_i}}g(\boldsymbol{x})\nabla_{{x_i}}f(\boldsymbol{x}) + g(\boldsymbol{x})\nabla_{{x_i}}^2f(\boldsymbol{x}).
\label{integration}
\end{aligned}
\end{equation} 
\end{footnotesize}

We consider this as a function of $x_i$ alone, all other variables being fixed. Then,
integrating over $x_i \in \mathbb{R}$, we get the equation:

\begin{footnotesize}
\begin{equation}
\begin{aligned}
 \lim_{M\to\infty}g({\boldsymbol{x}})\nabla_{{x_i}}f({\boldsymbol{x}})|_{-\boldsymbol{M}_i}^{+\boldsymbol{M}_i}&=\int_{-\infty}^{+\infty} \nabla_{{x_i}}f(\boldsymbol{x})\nabla_{{x_i}}g(\boldsymbol{x}) d{x_i}\\
 &+\int_{-\infty}^{+\infty}g(\boldsymbol{x})\nabla_{{x_i}}^2 f(\boldsymbol{x}) dx_i,
\label{integration_2}
\end{aligned}
\end{equation} 
\end{footnotesize}
which can prove the Eq. (\ref{integration_part}). ``+$\boldsymbol{M}_i$" represents the vector $[x_1,...,x_{i-1},+M,x_{i+1},...,x_n]$. ``-$\boldsymbol{M}_i$" represents the vector $[x_1,...,x_{i-1},-M,x_{i+1},...,x_n]$.

\section*{Detailed Derivation of the Equation (\ref{ACGalignment_third})}
\begin{footnotesize}
\begin{equation}
\begin{aligned}
     &\int_{-\infty}^{+\infty}\hspace{-0.2cm} p_D(y) dy \int_{\boldsymbol{x} \in \mathbb{R}^n}   (\nabla_{\boldsymbol{x}}\log p_{\theta}(y|\boldsymbol{x})^{\rm{T}}\cdot\nabla_{\boldsymbol{x}}\log p_D({\boldsymbol{x}}|y)) p_D(\boldsymbol{x}|y) d\boldsymbol{x} \\
    &\overset{\text{(I)}}= \int_{-\infty}^{+\infty}\hspace{-0.2cm} p_D(y) dy\int_{\boldsymbol{x} \in \mathbb{R}^n}   (\nabla_{\boldsymbol{x}}\log p_{\theta}(y|\boldsymbol{x})^{\rm{T}}\cdot\nabla_{\boldsymbol{x}} p_D({\boldsymbol{x}}|y))  d\boldsymbol{x} \\
    &\overset{\text{(II)}}= \int_{-\infty}^{+\infty}\hspace{-0.2cm} p_D(y) dy  \sum_{i=1}^{n} \int_{\boldsymbol{x} \in \mathbb{R}^n} \nabla_{{x_i}}\log p_{\theta}(y|\boldsymbol{x})\nabla_{{x_i}} p_D({\boldsymbol{x}}|y) d {\boldsymbol{x}}\\
    &= \int_{-\infty}^{+\infty}\hspace{-0.3cm} p_D(y) dy  \sum_{i=1}^{n} \int_{\tilde{\boldsymbol{x}}_i \in \mathbb{R}^{n-1}} [ \int_{-\infty}^{+\infty}\hspace{-0.35cm} \nabla_{{x_i}}\log p_{\theta}(y|\boldsymbol{x})\nabla_{{x_i}} p_D({\boldsymbol{x}}|y) d {x_i}] d{\tilde{\boldsymbol{x}}_i}\\    
    &\overset{\text{(III)}}=\hspace{-0.1cm}\int_{-\infty}^{+\infty}\hspace{-0.35cm} p_D(y) dy \sum_{i=1}^{n} \int_{\tilde{\boldsymbol{x}}_i \in \mathbb{R}^{n-1}} [\lim_{M\to\infty} p_D(\boldsymbol{x}|y) \nabla_{x_i}\log p_{\theta}(y|\boldsymbol{x})|_{-\boldsymbol{M}_i}^{+\boldsymbol{M}_i}]d{\tilde{\boldsymbol{x}}_i} \\
    &\quad - \int_{-\infty}^{+\infty}\hspace{-0.2cm} p_D(y) dy \sum_{i=1}^{n} \int_{\tilde{\boldsymbol{x}}_i \in \mathbb{R}^{n-1}} [\int_{-\infty}^{+\infty}\hspace{-0.2cm} p_D(\boldsymbol{x}|y) \nabla^2_{x_i} \log p_{\theta}(y|\boldsymbol{x}) d{x_i}]d{\tilde{\boldsymbol{x}}_i}\\
    &=\hspace{-0.1cm}\int_{-\infty}^{+\infty}\hspace{-0.35cm} p_D(y) dy \sum_{i=1}^{n} \int_{\tilde{\boldsymbol{x}}_i \in \mathbb{R}^{n-1}} [\lim_{M\to\infty} p_D(\boldsymbol{x}|y) \nabla_{x_i}\log p_{\theta}(y|\boldsymbol{x})|_{-\boldsymbol{M}_i}^{+\boldsymbol{M}_i}]d{\tilde{\boldsymbol{x}}_i} \\
    &\quad - \int_{-\infty}^{+\infty}\hspace{-0.2cm} p_D(y) dy \sum_{i=1}^{n} \int_{\boldsymbol{x} \in \mathbb{R}^{n}} [ p_D(\boldsymbol{x}|y) \nabla^2_{x_i} \log p_{\theta}(y|\boldsymbol{x})] d{\boldsymbol{x}}\\    
    &=\hspace{-0.1cm}\int_{-\infty}^{+\infty}\hspace{-0.35cm} p_D(y) dy \sum_{i=1}^{n} \int_{\tilde{\boldsymbol{x}}_i \in \mathbb{R}^{n-1}} [\lim_{M\to\infty} p_D(\boldsymbol{x}|y) \nabla_{x_i}\log p_{\theta}(y|\boldsymbol{x})|_{-\boldsymbol{M}_i}^{+\boldsymbol{M}_i}]d{\tilde{\boldsymbol{x}}_i} \\
    &\quad - \mathbb{E}_{p_D(y)}\mathbb{E}_{p_D({\boldsymbol{x}}|y)} [ {\rm{tr}} (\nabla^2_{\boldsymbol{x}} \log p_{\theta}(y|{\boldsymbol{x}})) ]\\
    &\overset{\text{(IV)}}=- \mathbb{E}_{p_D(y)}\mathbb{E}_{p_D({\boldsymbol{x}}|y)} \left[ {\rm{tr}} (\nabla^2_{\boldsymbol{x}} \log p_{\theta}(y|{\boldsymbol{x}}))\right],
\label{ACGalignment_third_appendix}
\end{aligned}
\end{equation}
\end{footnotesize}
where $\nabla^2_{\boldsymbol{x}}$ denotes the Hessian with respect to $\boldsymbol{x}$. ``+$\boldsymbol{M}_i$" represents the vector $[x_1,...,x_{i-1},+M,x_{i+1},...,x_n]$. ``-$\boldsymbol{M}_i$" represents the vector $[x_1,...,x_{i-1},-M,x_{i+1},...,x_n]$. $\boldsymbol{x}=[x_1,...,x_n]$ is an n-dimensional vector. $\tilde{\boldsymbol{x}}_i=[x_1,...,x_{i-1},x_{i+1},...,x_n]$.

We use the formula: $\nabla_x \log f(x) = f(x)^{-1}\nabla_x f(x)$ for equality (I). 
{In equality (I), $\nabla_{\boldsymbol{x}}\log p_{\theta}(y|\boldsymbol{x})^{\rm{T}}$ and $\nabla_{\boldsymbol{x}}\log p_D({\boldsymbol{x}}|y)$ are n-dimensional vectors, and their product result is a scalar.}
We use the formula: $\boldsymbol{u}^T \cdot \boldsymbol{v} = \sum_{i=1}^{n} u_i v_i$ for equality (II), where n represents the dimension of the data. 
As for equality (III), we use the integration by parts formula.
The equality (IV) holds for that we assume $p_D({\boldsymbol{x}}|y) \rightarrow 0$ {when $||\boldsymbol{x}||_2 \rightarrow \infty$.}

}

 
 {\small
\bibliographystyle{IEEEtranN}

\bibliography{newIEEEabrv}
}

\vfill

\end{document}